\documentclass[11pt]{article}

\usepackage[final]{latex/acl}

\usepackage{times}
\usepackage{latexsym}

\usepackage[T1]{fontenc}

\usepackage[utf8]{inputenc}

\usepackage{microtype}

\usepackage{inconsolata}

\usepackage{graphicx}

\usepackage{amsmath}

\usepackage{multirow}
\usepackage{booktabs} 

\usepackage{graphicx}
\usepackage{subcaption}

\usepackage{float}

\usepackage{longtable}
\usepackage{booktabs} 
\usepackage[table]{xcolor}
\usepackage{array} 

\usepackage{threeparttable}

\usepackage{tabularx}

\usepackage{arydshln}

\usepackage{longtable}

\usepackage{cleveref}

\usepackage[nottoc]{tocbibind}
\usepackage{minitoc}
\usepackage{bbm}

\newcommand{\sun}[1]{{\color{red}{\small\bf\sf [Sun: #1]}}}
\newcommand{\chun}[1]{{\color{brown}{\small\bf\sf [lichun: #1]}}}

\newcommand{\rf}[1]{} 
\newcommand{\Skip}[1]{}

\newcommand{\ie}{\textit{i}.\textit{e}.,\ }
\newcommand{\eg}{\textit{e}.\textit{g}.,\ }

\newcommand{\dotieconcat}[2]{
  \text{\raisebox{.8ex}{$\smallfrown$}}%
}

\makeatletter
\newcommand\dslfontsize{\@setfontsize\dslfontsize\@viipt\@viiipt}
\renewcommand\scriptsize{\@setfontsize\subfigcap{7}{8}}%
\makeatother

\newcommand\blfootnote[1]{%
  \begingroup
  \begin{NoHyper}%
  \renewcommand\thefootnote{}\footnote{#1}%
  \addtocounter{footnote}{-1}%
  \end{NoHyper}%
  \endgroup
}

\usepackage{color}
\usepackage{xcolor}
\definecolor{codegray}{rgb}{0.5,0.5,0.5}

\title{BILLY\hspace{0.05cm}\raisebox{-.15\height}{\includegraphics[scale=0.08]{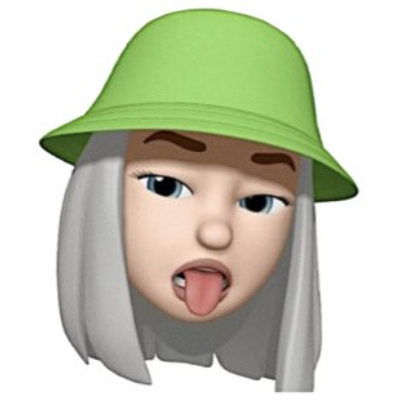}}: Steering Large Language Models via Merging Persona Vectors for Creative Generation}

\author{
 \textbf{Tsung-Min Pai\textsuperscript{1}}
 \quad
 \textbf{Jui-I Wang\textsuperscript{2}} 
 \quad
 \textbf{Li-Chun Lu\textsuperscript{3}}
 \quad
 \\
 \textbf{Shao-Hua Sun\textsuperscript{1,3}}
 \quad
 \textbf{Hung-Yi Lee\textsuperscript{1,3}}
 \quad
 \textbf{Kai-Wei Chang\textsuperscript{4}}
\\
 \textsuperscript{1}Department of Electrical Engineering, National Taiwan University\\
 \textsuperscript{2}Department of Computer Science \& Information Engineering, National Taiwan University\\
 \textsuperscript{3}Graduate Institute of Communication Engineering, National Taiwan University\\
 \textsuperscript{4}CSAIL, Massachusetts Institute of Technology
\\
 \small{
   \textbf{Correspondence to:}
   Tsung-Min Pai <\href{mailto:b09602017@ntu.edu.tw}{b09602017@ntu.edu.tw}>,
   Kai-Wei Chang <\href{mailto:kwchang@mit.edu}{kwchang@mit.edu}>
 }
}

\begin{document}
\doparttoc 
\faketableofcontents 

\maketitle

\begin{abstract}

Multi-LLM systems enhance the creativity of large language models by simulating human collective intelligence but suffer from significant drawbacks, such as high computational costs and inference latency.
To address these limitations, we propose \textbf{BILLY} (\textbf{B}lend\textbf{I}ng persona vectors for \textbf{L}arge \textbf{L}anguage model creativit\textbf{Y}), a training-free framework that captures the benefits of multi-LLM collaboration, \ie inducing diverse perspectives and specialized expertise, within a single model. 
BILLY operates by extracting and blending multiple distinct persona vectors directly in the model's activation space. We steer the model's generation process with this merged vector while inference, enabling multi-perspective output without explicit multi-LLM communication. 
Our experiments across creativity-oriented benchmarks demonstrate that BILLY surpasses single model prompting and traditional multi-LLM approaches, while substantially reducing inference time and computational costs. 
Our analyses further reveal that distinct persona vectors can be blended to achieve both effective control over complementary aspects of generation and greater interpretability.
Our project page and codes are available at \href{https://bai1026.github.io/LLM_Persona/}{Website} and \href{https://github.com/Bai1026/LLM_Persona/tree/main?tab=readme-ov-file}{GitHub}.
\end{abstract}

\section{Introduction}
Creativity is widely recognized as a cornerstone of human progress, driving innovation across domains and enabling major scientific breakthroughs~\cite{feist1998meta, simonton2004creativity}.
Extending this perspective, recent research~\cite{DBLP:journals/corr/abs-2304-00008, MAS_survey, lu2026rethinking} has explored the creativity of large language models (LLMs), viewing them as promising tools for applications such as story writing~\cite{gomez-rodriguez-williams-2023-confederacy, hollmwood}, design ideation~\cite{MAS_design, hou2024c2ideas, hung2025simtube}, and scientific discovery~\cite{yang2024large, si2025can}, thereby augmenting human problem-solving and imagination~\cite{zhang2023bootstrap, Co_GPT, MAS_Coquest, liu2025synthesizing}.

\begin{figure}[t]
    \centering
    \includegraphics[width=0.99\linewidth]{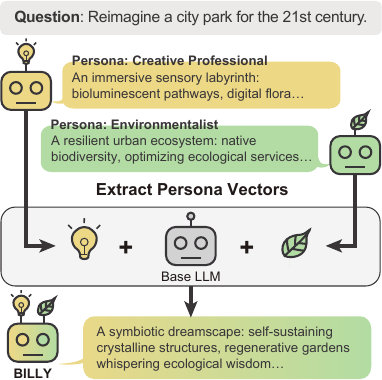}
    \caption{
    \textbf{BILLY} (\textbf{B}lend\textbf{I}ng persona vectors for \textbf{L}arge \textbf{L}anguage model creativit\textbf{Y}).
    To enhance the creativity of a single LLM, we extract and fuse the persona vectors of a \texttt{Creative Professional} and an \texttt{Environmentalist}, steering a base model by this composite vector to generate outputs based on both domains.
    }
    \label{fig:intro_figure}
\end{figure}

Recent studies highlight the potential of the \emph{multi-LLM} paradigm~\cite{guo2024large, llm_discussion, MAS_survey}, which aims to simulate human collective intelligence~\cite{leimeister2010collective} by engaging multiple LLMs in iterative discussion to arrive at more comprehensive and well-balanced solutions~\cite{tran2025multi}. This paradigm allows systems 
to generate a broader range of ideas beyond a single model can reach~\cite{tran2025multi}. In the context of creativity, these frameworks often assign diverse roles to LLMs and employ structured, multi-round interactions~\cite{llm_discussion, MAS_survey, many-heads, MAS_debate}.
\blfootnote{Our method, BILLY, takes its name from \textit{The Minds of Billy Milligan} by Daniel Keyes—but the emoji in the title \raisebox{-.25\height}{\includegraphics[scale=0.05]{Figures/billie.png}}
nods to “Billie” Eilish instead.}
For example, one LLM may act as a creative professional while another serves as an environmentalist, enabling the system to explore and integrate multiple perspectives. Through such iterative exchanges, the models collaboratively converge on creative solutions that balance innovation with sustainability. 

Despite these advantages, multi-LLM frameworks also face a major challenge: the substantial computational cost and increased inference time associated with multi-LLM coordination~\cite{tran2025multi}. The iterative nature of multi-LLM dialogue substantially increases the system's input and output tokens, as models must consume and produce multiple rounds of text while processing other models’ outputs. This multi-round exchange makes the process computationally costly and also increases inference latency~\cite{MAS_survey, MAS_cost, MAS_flatten}, limiting scalability in real-world settings. Moreover, multi-LLM frameworks can exhibit sociological drawbacks such as \emph{process loss}~\cite{steiner1972group}, where the collective outcome falls short of the sum of individual contributions due to communication friction or coordination inefficiencies~\cite{discussion_loss, cognitive_distnace}. Together, these issues expose a fundamental trade-off: while multi-LLM systems could potentially enhance diversity and creative potential, they do so at the expense of efficiency.

We therefore ask: \emph{Without relying on multiple models carrying on a multi-round discussion, can a single LLM simulate diverse perspectives in an efficient manner?} A natural way to explore this idea is through prompting, that is, asking the model to simultaneously adopt different roles and viewpoints. For example, one might prompt an LLM to act as both a creative professional and an environmentalist when producing a solution that balances originality with sustainability.
However, this approach depends heavily on the model’s ability to integrate and coordinate multiple roles and perspectives within a single model and single response.
In practice, an LLM may handle each persona separately but struggle to integrate them coherently in a single pass, often due to limitations in capability or the coverage of the training data~\cite{ActAdd}.

To address these challenges while retaining the creative advantages of multi-LLM interaction, we propose \textbf{BILLY}(\textbf{B}lend\textbf{I}ng persona vectors for \textbf{L}arge \textbf{L}anguage model creativit\textbf{Y}) as illustrated in \Cref{fig:intro_figure}. Inspired by recent advances in extracting \emph{persona vectors}~\cite{persona_vec} and steering model behavior through \emph{activation engineering}~\cite{ActAdd, CAA}, BILLY extends these ideas by merging multiple persona vectors within a single LLM. This enables the model to produce multi-perspective, creative responses without resorting to computationally heavy multi-LLM frameworks. Notably, BILLY is training-free and empirically effective and efficient—it produces creative outputs that consistently reflect the targeted personas. Our experiments show that BILLY achieves higher creativity than single-model prompting or multi-LLM methods while being substantially more cost-efficient.
We summarize our contributions as follows:

\begin{itemize}
    \item \textbf{Enhanced Creativity.} We propose the framework BILLY, which blends multiple persona vectors within a single LLM. It enables the generation of diverse, multi-perspective, and creative responses, surpassing both single-model prompting and multi-LLM approaches on creativity-oriented benchmarks.  

    \item \textbf{Efficiency and Simplicity.} BILLY is entirely \emph{training-free}, requiring no additional fine-tuning or multi-LLM communication. It achieves comparable or superior creativity with substantially lower computational and token costs.  

    \item \textbf{Interpretability via Persona Vectors.} Operating directly in the latent activation space, BILLY offers an interpretable mechanism for understanding and steering creativity, allowing fine-grained and transparent control through persona blending.  
\end{itemize}

\section{Related Work} 




\subsection{LLM Creativity}
Enhancing creativity in LLMs has been a growing research focus. One common approach involves role-playing prompts, which guide models to adopt specific creative  personas~\cite{chen2024personapersonalizationsurveyroleplaying}. Another works~\cite{llm_discussion, hollmwood, many-heads} explore multi-agent collaboration frameworks, where multiple agents simulate brainstorming sessions by proposing and critiquing ideas to achieve more sophisticated creative outcomes. Comprehensive reviews of this research direction can be found in~\citet{Chakrabarty2023ArtOA, gomez-rodriguez-williams-2023-confederacy}. 

\subsection{Multi-LLM Collaboration}
Multi-agent collaboration has emerged as a powerful paradigm for enhancing LLM applications \cite{Du2023ImprovingFA, chan2024chateval, Liu2023DynamicLN, sun2023corex}. AutoGen~\cite{Wu2023AutoGenEN} enables next-generation applications through structured conversations, while LLM Discussion~\cite{llm_discussion} and HOLLMWOOD~\cite{hollmwood} employ role-play and dialogue to stimulate more creative outputs. Beyond creativity, researchers have explored emergent cognitive synergy through multi-persona self-collaboration~\cite{wang-etal-2024-unleashing} and applied divergent-agent systems to domain-specific tasks such as electronic design automation~\cite{wu-etal-2025-divergent}. While infrastructures like AgentVerse~\cite{AgentVerse} and MetaGPT~\cite{hong2024metagpt} provide frameworks for coordination, these collaborations can suffer from high computational costs, instability, and inefficiencies~\cite{cemri2025multiagentllmsystemsfail, han2025llmmultiagentsystemschallenges}. The complexity of managing communication and coordination increases with the number of agents, potentially leading to bottlenecks and unpredictable behaviors.

\subsection{Activation Steering}
Activation steering is a technology that modifies a model’s internal activations along specific directions to control or influence its behavior~\cite{RepE, ActAdd}. This is achieved by first identifying a steering vector~\citep{liang2026adaptive}, which represents the desired behavioral shift. This vector is typically obtained using contrastive methods, where the model's activation differences are calculated from pairs of inputs that elicit opposite behaviors (e.g., a truthful vs. a deceptive response). During inference, this steering vector is added to the activations at a specific layer, effectively nudging the model's internal state toward the intended outcome.

A key advantage of this method is that it manipulates activations only at inference time without altering the model's underlying weights. This distinguishes it from techniques like model merging, which operates directly on model parameters to integrate capabilities from multiple fine-tuned models~\cite{yang2024modelmergingllmsmllms, sun2025personalityvectormodulatingpersonality, jang2023personalizedsoupspersonalizedlarge}, and offers a more lightweight, interpretable, and fine-grained control over model behavior. This technique has been applied across domains such as enhancing truthfulness~\cite{ITI, CAA}, improving safety~\cite{zhang-etal-2025-personalized, ActAdd}, and increasing transparency and interpretability~\cite{RepE, mean-centring}. Recent works have also leveraged activation steering for persona-related applications~\cite{mean-centring, ActAdd, RepE}. For example, Persona Vectors~\cite{persona_vec} identify activation directions corresponding to personalities for controllable character traits, while bi-directional preference optimization~\cite{BIPO} enables versatile steering vectors for fine-grained persona control. 

\section{Approach: BILLY}
\label{sec:approach}
Our approach, \textbf{BILLY}, \textbf{B}lend\textbf{I}ng persona vectors for \textbf{L}arge \textbf{L}anguage model creativit\textbf{Y}, is designed to overcome the limitations of prompting methods \cite{ActAdd, hollmwood, MAS_debate, hung2025simtube} and the prohibitive costs of interactive multi-agent frameworks \cite{many-heads, ICCRI, llm_discussion}.
We introduce a novel paradigm that shifts the control of the model from an external, conversational system to direct, internal representational steering to improve LLM creativity. The core of our method is a three-stage process: (1) extracting distinct persona vectors through a contrastive method, (2) fusing these vectors offline into a single composite steering vector, and (3) applying this vector during inference to guide a single model.

\subsection{Extracting Persona Vectors via Contrastive Activation}
\label{subsec:extraction}

A \textit{persona vector} \citep{persona_vec} represents the characteristics of a specific persona (e.g., evil, humorous) as a directional vector within the model's activation space \citep{RepE, wu2025axbench, CAA}, capturing the \textit{shift} in the model's activation states when it adopts a persona compared to when it responds neutrally. 
By adding such a vector during inference, we can steer the model's behavior to align with the desired persona.

Building upon the pipeline proposed by~\citet{persona_vec} for extracting persona vectors, we create a customized dataset designed to capture nuanced and specialized personas. This dataset enables more effective handling of creativity-oriented tasks and ensures a fair comparison with multi-LLM systems~\citep{llm_discussion, hollmwood, ICCRI}.

Given a persona $P$, we curate two distinct sets of model responses, \textbf{positive-expression set} $D_P^+$ and \textbf{negative-expression set} $D_P^-$, as follows: (1) design contrastive system prompts (positive vs. negative), (2) employ an LLM judge to score the alignment of each response with the corresponding trait, and (3) filter the responses with a threshold to ensure a clear distinction between the two corpora.

After the process, $D_P^+$ contains responses that exhibit the persona's traits, and $D_P^-$ contains baseline responses that lack these specific traits.

For any given response $\mathbf{x}$ from these sets, we extract its token-averaged residual stream activation at a specific layer $l$, which we denote as $\vec{a}^{(l)}(\mathbf{x})$. The persona vector for persona $P$ at layer $l$, denoted by $\vec{v}_P^{(l)}$, is then defined as the \textbf{difference between the mean activation vectors} of the positive and neutral sets:
\begin{equation}
\label{eq:persona_vector_definition_sum}
\vec{v}_P^{(l)} = \frac{1}{|D_P^+|} \sum_{\mathbf{x} \in D_P^+} \vec{a}^{(l)}(\mathbf{x}) - \frac{1}{|D_P^-|} \sum_{\mathbf{x} \in D_P^-} \vec{a}^{(l)}(\mathbf{x})
\end{equation}
This resulting vector, $\vec{v}_P^{(l)}$, captures the directional \textbf{shift} in the activation space required to induce the persona $P$. This procedure is repeated for all layers, generating a unique candidate vector for each layer. We present the detailed extraction process in \Cref{app:persona_vector_extraction}.

\subsection{Offline Fusion for a Composite Vector}
\label{subsec:fusion}
Unlike multi-LLM systems that require costly online interaction between agents, our approach fuses these perspectives in an offline step. Given a set of $N$ extracted persona vectors $\{\vec{v}_{1}, \vec{v}_{2}, \dots, \vec{v}_{N}\}$ that we wish to combine, we compute a single composite steering vector, $\vec{v}_{\text{merged}}$, by taking their average:
\begin{equation}
\vec{v}_{\text{merged}}^{(l)} = \frac{1}{N} \sum_{i=1}^{N} \vec{v}_{i}^{(l)}
\label{eq:merge_vector}
\end{equation}
This resulting $\vec{v}_{\text{merged}}$ represents a multi-faceted perspective in the model's activation space. This one-time calculation yields a reusable vector that encapsulates the essence of a multi-personality.

\subsection{Inference-time Steering with Composite Persona Vector}
\label{subsec:steering}
The final stage applies the composite vector to steer the model's generation process. During a standard forward pass, when the model computes the activations $\vec{a}_{\text{original}}^{(l)}$ at a chosen steering layer $l$, we intervene by adding our composite vector, scaled by a coefficient $\alpha$:

\begin{equation}
    \vec{a}_{\text{steered}}^{(l)} = \vec{a}_{\text{original}}^{(l)} + \alpha \cdot \vec{v}_{\text{merged}}^{(l)}
\label{eq:steering}
\end{equation}

This simple addition acts as a directional push in the activation space, steering the model's subsequent computations and thus its final output towards a state that aligns with the fused perspectives of the multiple personas.
Crucially, this steering process requires only a single additive operation during each inference step and involves no additional training. 
Specifically, layer 20 and coefficient 2.0 in our experiments, following \citet{persona_vec}, and we also conduct a study of steering layers and coefficients in Appendix~\ref{appen:choice}.


\section{Experiment}

\subsection{Benchmarks}

To evaluate the creativity of large language models (LLMs), we adopt a benchmark suite originally derived from human creativity tests, Wallach-Kogan Creativity Tests~\cite{wkct_test}. Rather than using the original tasks directly, we build upon a set of augmented benchmarks previously developed by~\citet{llm_discussion}, which expanded the original tasks using GPT-4 to generate additional items. This makes our evaluation based on a pre-augmented dataset, ensuring a broader and more diverse set of tasks while maintaining alignment with established creativity tests.

The benchmark set includes four benchmarks: \textsc{AUT} asks for practical and innovative uses for everyday objects; \textsc{INSTANCES} requires listing items according to specific criteria; \textsc{SIMILARITIES} involves identifying connections or similarities between items; and \textsc{SCIENTIFIC} focuses on producing creative solutions to scientific problems.
\Cref{table:dataset_sample} provides descriptions and examples from these benchmarks, with more detailed settings provided in the \Cref{appB:prompt}.
~




\begin{table*}[t!]
\centering
\small
\setlength{\tabcolsep}{6pt}
\begin{tabular}{p{1.6cm} p{7.0cm} p{5.5cm}}
\toprule
\textbf{Benchmark} & \textbf{Description} & \textbf{Sample Task} \\
\midrule
\textsc{AUT} &
Evaluates divergent thinking by requiring the generation of numerous unconventional applications for an object. &
What are some creative uses for a mug? \\

\midrule
\textsc{Instances} &
Measures the ability to produce a diverse set of examples that satisfy a given categorical or physical property. &
Name 5 the square things you can think of. \\
\midrule

\textsc{Similarities} &
Assesses associative creativity by challenging participants to identify and articulate non-obvious connections between two distinct concepts. &
Tell me 5 ways in which a brick and a stone are alike. \\

\midrule
\textsc{Scientific} &
Probes creative problem-solving within a scientific framework, encompassing hypothesis generation, experimental design, and technological enhancement. &
Find different scientific uses for a spoon (e.g., demonstrating reflection or heat conduction). \\
\bottomrule
\end{tabular}
\caption{\label{table:dataset_sample}\textbf{Creativity Benchmarks Overview.} The first three benchmarks (\textsc{AUT}, \textsc{Instances}, \textsc{Similarities}) target general cognitive creativity, while \textsc{Scientific} assesses creative aptitude in a structured, domain-specific context.}
\end{table*}


\subsection{Baselines}
We include both single agent prompting methods and multi-agent systems as our baselines. 
\Cref{appB:prompt} shows the prompts we use in the baseline methods.
Our baselines are as follows:

\noindent \textbf{(1) Single Agent (SA)}: A single LLM prompted to respond creatively, with the temperature set to $0.7$,
    which is a commonly used value.

\noindent \textbf{(2) SA with High-temperature Decoding, SA ($T=1.0$)}: The temperature $T$ is increased to $1.0$ to stimulate higher levels of diversity. By allowing the model to explore a broader range of possible outputs, this setting encourages more diverse responses. We include this configuration to test whether greater diversity in generation can lead to more creative outcomes, though it may also introduce greater variability in quality~\cite{peeperkorn2024temperaturecreativityparameterlarge}.

\noindent \textbf{(3) SA with Multi-Role Prompt (SA-MRP)}: In addition to the standard single agent baseline, we design a multi-role prompting variant. The prompt here is extended with explicit persona instructions. The model is asked to respond from multiple professional perspectives, such as environmentalist, creative professional, and futurist, each with distinct expertise styles. This approach is included as a strong baseline to determine the performance limits of enhancing creativity through prompt engineering alone. By comparing our proposed BILLY against this advanced prompting technique, we can more clearly demonstrate the advantages of directly manipulating the model's internal activations over purely prompt-level modifications.

\noindent \textbf{(4) LLM Discussion}: A multi-LLM framework proposed by \citet{llm_discussion}, which organizes multiple LLM agents into a structured three-phase process—initiation, discussion, and convergence—with each agent role-played under distinct personas to diversify perspectives. Agents exchange ideas over several rounds and then consolidate them into final outputs. Our implementation follows the original setup, including role assignments and discussion prompts.

\subsection{Evaluation}
To quantify the creative performance of each method, we adopt two primary metrics from the Torrance Tests of Creative Thinking (TTCT; \citealp{TTCT}) and perform both LLM-based evaluation (GPT-4o-mini) and human evaluation. The TTCT is a widely respected instrument in psychological research, valued for its reliability in measuring creative output~\cite{TTCT_report}. From its framework, we focus on the two key dimensions most relevant to the quality of generated ideas. We evaluate the whole QA pair with these two metrics.

\paragraph{Metrics:}
The specific metrics employed are as follows:
\begin{itemize}
    \item \textbf{Originality}: This metric assesses the statistical rarity or unconventionality of a response. Ideas are scored based on their novelty and divergence from common or obvious answers, while still maintaining relevance to the prompt.
    \item \textbf{Elaboration}: This metric measures the level of detail and supportive information within a response. It evaluates the ability to expand upon a core idea, adding depth, context, and specific details to enrich the initial concept.
\end{itemize}

The TTCT framework includes two other metrics: fluency and flexibility. However, as discussed in \citet{llm_discussion}, these metrics are not suitable for assessing LLM creativity. The authors point out that while fluency and flexibility pose significant challenges for human subjects, they are trivial for LLMs, which can generate responses in any desired quantity or variety on demand. Our preliminary study confirms this finding. Therefore, since these metrics do not serve as meaningful differentiators for LLM creativity, our evaluation will proceed using only originality and elaboration.

Both human evaluation and LLM-based evaluation follow the same TTCT-based scoring rubric, as presented in \Cref{appB:prompt}, to ensure consistency across evaluation settings. 


\definecolor{1}{HTML}{EAD1F7} 
\definecolor{2}{HTML}{DDEBF7} 

\makeatletter
\newcommand{\hlbox}[2][1]{%
  \begingroup
  \setlength{\fboxsep}{1pt}
  \colorbox{#1}{\vphantom{Ay}\smash{\strut #2}}%
  \endgroup
}
\makeatother

\newcommand{\hlone}[1]{\hlbox[1]{#1}}
\newcommand{\hltwo}[1]{\hlbox[2]{#1}}

\begin{table*}[h!]
\centering
\begin{threeparttable}
\small

\newcommand{\pvm}{BILLY (Ours)}

\newcommand{\llmd}{LLM Discussion}
\newcommand{\savanilla}{SA}
\newcommand{\sarole}{SA-MRP}
\newcommand{\satemp}{SA (T=1.0)}

\scalebox{0.92}{\begin{tabular}{llcccccc}
\toprule
\multirow{2}{*}{\textbf{Task}} & \multirow{2}{*}{\textbf{Method}} &
\multicolumn{2}{c}{\textbf{Qwen-2.5-7B-Instruct}} &
\multicolumn{2}{c}{\textbf{Llama-3.1-8B-Instruct}} &
\multicolumn{2}{c}{\textbf{Gemma-3-4B-it}} \\
\cmidrule(lr){3-4} \cmidrule(lr){5-6} \cmidrule(lr){7-8}
 &  & Originality & Elaboration & Originality & Elaboration & Originality & Elaboration \\
\midrule

\multirow{5}{*}{\textsc{AUT}}
 & \savanilla & 4.03±0.55 & 4.84±0.37 & 3.73±0.59 & 4.69±0.48 & 4.79±0.16 & 4.94±0.26 \\
 & \satemp    & 4.15±0.51 & 4.86±0.36 & 3.75±0.63 & 4.60±0.52 & 4.78±0.42 & 4.93±0.25 \\
 & \sarole    & 4.07±0.44 & \cellcolor{1}{4.91±0.29} & 4.05±0.47 & 4.70±0.49 & \cellcolor{2}{4.96±0.19} & \cellcolor{1}{4.97±1.11} \\
 & \llmd      & \cellcolor{2}{4.24±0.54} & \cellcolor{2}{4.88±0.33} & \cellcolor{2}{4.21±0.59} & \cellcolor{1}{4.85±0.39} & 3.70±1.25\tnote{*} & 4.03±1.29 \\
 & \pvm       & \cellcolor{1}{4.71±0.45} & 4.86±0.37 & \cellcolor{1}{4.38±0.55} & \cellcolor{2}{4.84±0.39} & \cellcolor{1}{4.99±0.08} & \cellcolor{2}{4.96±0.21} \\
 \midrule

\multirow{5}{*}{\textsc{Instances}}
 & \savanilla & 3.35±0.77 & 3.96±0.62 & 2.57±0.70 & 3.63±0.62 & \cellcolor{2}{4.85±0.35} & 4.82±0.40 \\
 & \satemp    & \cellcolor{2}{3.49±0.74} & 3.97±0.60 & 2.76±0.73 & 3.81±0.61 & 4.64±0.52 & \cellcolor{1}{4.87±0.34} \\
 & \sarole    & 3.47±0.56 & \cellcolor{2}{4.54±0.61} & 3.10±0.63 & \cellcolor{2}{4.35±0.58} & 4.83±0.38 & 4.84±0.37 \\
 & \llmd      & 3.29±0.76 & \cellcolor{1}{4.55±0.55} & \cellcolor{2}{3.46±0.73} & 4.08±0.66 & 3.72±1.03\tnote{*} & 4.06±1.10 \\
 & \pvm       & \cellcolor{1}{4.57±0.55} & 4.41±0.71 & \cellcolor{1}{4.53±0.60} & \cellcolor{1}{4.78±0.43} & \cellcolor{1}{4.96±0.20} & \cellcolor{2}{4.86±0.34} \\
\midrule

\multirow{5}{*}{\textsc{Scientific}}
 & \savanilla & 3.63±0.65 & \cellcolor{2}{4.67±0.47} & 3.65±0.62 & \cellcolor{2}{4.63±0.50} & \cellcolor{1}{4.96±0.20} & 4.69±0.49 \\
 & \satemp    & \cellcolor{2}{3.74±0.64} & \cellcolor{1}{4.76±0.44} & \cellcolor{2}{3.70±0.70} & \cellcolor{1}{4.68±0.48} & 4.51±0.51 & \cellcolor{1}{4.82±0.38} \\
 & \sarole    & 3.65±0.54 & 4.54±0.54 & 3.54±0.58 & 4.58±0.53 & \cellcolor{1}{4.96±0.20} & \cellcolor{2}{4.75±0.44} \\
 & \llmd      & 3.45±0.66 & 4.64±0.58 & 3.24±0.70 & 4.26±0.65 & 3.76±1.03\tnote{*} &	4.28±0.97 \\
 & \pvm       & \cellcolor{1}{4.48±0.40} & 4.32±0.67 & \cellcolor{1}{4.48±0.40} & 4.51±0.46 & \cellcolor{2}{4.94±0.24} & 4.68±0.50 \\
\midrule

\multirow{5}{*}{\textsc{Similarities}}
 & \savanilla & 2.50±0.48 & 4.68±0.33 & 3.12±0.47 & \cellcolor{2}{4.73±0.28} & \cellcolor{1}{4.97±0.24} & 4.75±0.46 \\
 & \satemp    & 2.62±0.68 & 4.67±0.47 & \cellcolor{2}{3.33±0.70} & \cellcolor{1}{4.75±0.47} & 4.41±0.51 & \cellcolor{1}{4.92±0.28} \\
 & \sarole    & 3.08±0.53 & \cellcolor{2}{4.81±0.41} & 3.01±0.51 & 4.63±0.48 & 4.93±0.26 & 4.76±0.44 \\
 & \llmd      & \cellcolor{2}{3.19±0.74} & \cellcolor{1}{4.84±0.37} & 2.91±0.74 & 4.37±0.61 & 3.70±0.95\tnote{*} &	4.17±0.95 \\
 & \pvm       & \cellcolor{1}{4.36±0.68} & 4.66±0.58 & \cellcolor{1}{4.39±0.68} & 4.68±0.50 & \cellcolor{2}{4.94±0.24} & \cellcolor{2}{4.82±0.40} \\
\bottomrule
\end{tabular}}
\end{threeparttable}
\caption{
\textbf{Evaluation Results.} We use Qwen-2.5-7B-Instruct, Llama-3.1-8B-Instruct, and Gemma-3-4B-it as our base models for steering and employ GPT-4o-mini as the LLM judge. 
The \hlone{highest} scores in each benchmark are highlighted in purple, while the \hltwo{second-highest} is in blue.
Across the four benchmarks, the Originality scores of our method BILLY surpass nearly all of the baselines. 
}
\label{tab:main_result}
\end{table*}

\subsection{Results}
\label{sec:resutls}
\subsubsection{LLM-based Evaluation}
Our primary experiments evaluate BILLY against several baselines by steering three distinct open-source models: Qwen-2.5-7B-Instruct, Llama-3.1-8B-Instruct, and Gemma-3-4B-it.
The main results, aggregated across all four creativity benchmarks, are presented in \Cref{tab:main_result}. 

The central finding of our work is that BILLY achieves a superior balance between \textbf{Originality} and \textbf{Elaboration}. Across four benchmarks, our method consistently outperforms all baseline methods in Originality while remaining competitive in terms of Elaboration.
This demonstrates the effectiveness of using internal representational control to elicit creative responses. For both Qwen and Llama models, BILLY surpasses the strong but costly LLM Discussion, as well as various single-agent configurations. 
Notably, while the SA-MRP occasionally provides a slight improvement over the SA, its performance is inconsistent, reinforcing our hypothesis that prompt-based control is inherently less reliable than direct vector steering.
To validate our findings, we conducted paired t-tests across our entire experimental matrix. The analysis reveals that BILLY's advantage is statistically significant (p-value < 0.05) in 74\% of all comparisons, and highly significant (p-value < 0.001) in 66\% of them. The majority of these significant improvements were observed on the Originality metric, reinforcing our central claim. Furthermore, we discuss the difference between linguistic diversity and true creativity in \Cref{app:diversity_vs_creativity}.

The Gemma-3-4B-it model, a smaller model, struggles with limitations in the long-form, multi-round LLM Discussion tasks, as its output tends to become unstable with the expanding context. We attribute its lower scores and high variance to this instability.
Notably, while BILLY still achieved the highest Originality score on this model in most cases, the SA baselines also exhibit exceptionally strong creative performance.

\subsubsection{Human Evaluation and Correlation}


\definecolor{1}{HTML}{EAD1F7} 
\definecolor{2}{HTML}{DDEBF7} 



\begin{table}[t]
\centering
\begin{threeparttable}
\small

\newcommand{\pvm}{BILLY (Ours)}
\newcommand{\llmd}{LLM Discussion}
\newcommand{\savanilla}{SA}

\scalebox{0.92}{
\begin{tabular}{p{1.6cm} l p{1.4cm} p{1.4cm}}
\toprule
\textbf{Task} & \textbf{Method} & \textbf{Originality} & \textbf{Elaboration} \\
\midrule

\multirow{3}{*}{\textsc{AUT}}
 & \savanilla & 3.11±1.21 & \cellcolor{2}{3.43±1.10} \\
 & \llmd      & \cellcolor{2}{3.55±1.07} & 2.78±1.09 \\
 & \pvm       & \cellcolor{1}{3.67±1.11} & \cellcolor{1}{4.08±0.89} \\
\midrule

\multirow{3}{*}{\textsc{Instances}}
 & \savanilla & 2.53±1.15 & 2.55±1.16 \\
 & \llmd      & \cellcolor{2}{3.45±1.19} & \cellcolor{2}{3.04±1.08} \\
 & \pvm       & \cellcolor{1}{3.75±1.38} & \cellcolor{1}{3.96±0.90} \\
\midrule

\multirow{3}{*}{\textsc{Scientific}}
 & \savanilla & 3.01±1.37 & \cellcolor{2}{3.86±1.07} \\
 & \llmd      & \cellcolor{2}{3.30±1.16} & 2.70±0.93 \\
 & \pvm       & \cellcolor{1}{3.47±1.18} & \cellcolor{1}{4.23±0.85} \\
\midrule

\multirow{3}{*}{\textsc{Similarities}}
 & \savanilla & \cellcolor{2}{3.15±1.40} & \cellcolor{2}{3.79±0.99} \\
 & \llmd      & 2.79±1.11 & 2.33±1.13 \\
 & \pvm       & \cellcolor{1}{3.75±1.09} & \cellcolor{1}{4.17±1.07} \\
\bottomrule
\end{tabular}
}

\caption{
\textbf{Human Evaluation Results.}
The \hlone{highest} scores are highlighted in purple, while the \hltwo{second-highest} are in blue.
Across all benchmarks, \pvm{} consistently achieves the strongest Originality and Elaboration scores.
}
\label{tab:human_eval}
\end{threeparttable}
\end{table}

\begin{table}[t]
\small
\centering
\begin{tabular}{lcc}
\toprule
\textbf{Correlation} & \textbf{Originality} & \textbf{Elaboration} \\
\midrule
Spearman & 0.73 & 0.43 \\
Pearson  & 0.66 & 0.40 \\
\bottomrule
\end{tabular}
\caption{\textbf{Correlation Between Human and LLM Judges.} Spearman and Pearson correlations between LLM-based evaluations and average  human evaluations for Originality and Elaboration.}
\label{tab:judges_correlation}
\end{table} 

Human evaluation is conducted on a subset of methods due to practical constraints. Specifically, we select three representative methods and evaluate them across all four tasks. For each task, three samples are randomly selected for human assessment. In total, we collect 132 evaluation scores contributed by 11 volunteer human evaluators.

The human evaluation results are shown in \Cref{tab:human_eval}. Overall, the trends are consistent with our LLM-based evaluation. Across all benchmarks, BILLY achieves the highest average scores in both Originality and Elaboration, outperforming the SA and the LLM Discussion method. BILLY shows particularly strong gains in Elaboration, indicating its ability to produce more detailed responses. Meanwhile, although LLM Discussion often improves originality over SA, its elaboration scores are consistently lower. These results suggest that direct activation steering provides a more effective and stable mechanism for eliciting creative responses.

We also calculated the Spearman's Rank Correlation Coefficient and Pearson Correlation Coefficient between our LLM-Judge and human raters, with the results show in \Cref{tab:judges_correlation}.
The results indicate a strong positive correlation for the Originality. While the correlation for the Elaboration score is more moderate, this aligns with observations reported in the LLM-Discussion \cite{llm_discussion}.

\section{Discussion \& Analysis}
\label{sec:discussion}
Our analysis goes beyond simple confirmation of efficacy to provide deeper insights into the mechanisms of BILLY. To dissect these mechanisms, our discussion is structured around four concepts: ($1$) The actual token cost and inference time among different methods.($2$) The relationship between the quantity of fused vectors and the resulting creative quality. ($3$) Qualitative nature of the generated text. ($4$) Layer-wise projection among different methods.
The experiments in this section adopt the Llama-3.1-8B-Instruct model, which we find sufficiently representative for our analysis.

\subsection{Token Cost and Inference Time}
\label{sec:cost}
\Cref{tab:time_and_cost_comparison} details the average inference latency, input and output token counts per query, and estimated token cost per $10000$ queries for each method.
The data clearly demonstrates BILLY's superior efficiency, reducing both token usage and associated estimated cost by \textbf{more than 95\%} compared to the LLM Discussion baseline. 
Notably, the amortized approach emerges as the most cost-effective solution, as its operational costs remain fixed regardless of query volume.
We discuss the amortized cost, which is the most evident advantage of BILLY, in \Cref{app:cost_amortized}, and \Cref{app:llm_discussion_log} also provides the complete chat log of a specific agent in LLM Discussion with a single question.

Practical efficiency of BILLY is evident not only in its token cost but also in its substantially shorter inference time.
We measured the wall-clock time required for each method to answer 100 \textsc{AUT} questions and calculated the average inference latency per query. As detailed in \Cref{tab:time_and_cost_comparison}, BILLY is approximately \textbf{25 times faster} than LLM Discussion. This dramatic reduction in latency suggests that our approach is a practical solution for real-time creative generation.


\begin{table}[H]
\centering
\small
\begin{threeparttable}
    \scalebox{0.9}{\begin{tabular}{lccc}
    \toprule
    \textbf{Method} & \textbf{Latency (s)} & \textbf{Token (In / Out)}\tnote{1} & \textbf{Cost (\$)}\tnote{2} \\
    \midrule
    SA & 23.9 & 22.3 / 407.1 & 0.25 \\
    SA-MRP & 36.8 & 221.2 / 861.1 & 0.56 \\
    LLM Discussion & 513 & 88853 / 12922.1 & 25.50 \\
    BILLY (Ours) & 19 & 62.2 / 475.6 & 0.30 \\
    \bottomrule
    \end{tabular}}
    \begin{tablenotes}
      \small
      \item[1] Token are per-query averages (n=10,000).
      \item[2] Cost is the estimated price for 10000 queries calculated\\
      based on \href{https://studio.nebius.com/}{Nebius AI Studio}, where the price is \$0.02 /\\ \$0.06 USD per million input / output tokens.
    \end{tablenotes}
\caption{\textbf{Inference Time and Token Cost.} BILLY demonstrates a significant reduction in both latency and token cost compared to LLM Discussion.}
\label{tab:time_and_cost_comparison}
\end{threeparttable}
\end{table}

\subsection{Analysis of Various Vector Compositions}
\label{subsec:ablation}
\begin{figure}[t]
    \centering
    \includegraphics[width=1\linewidth]{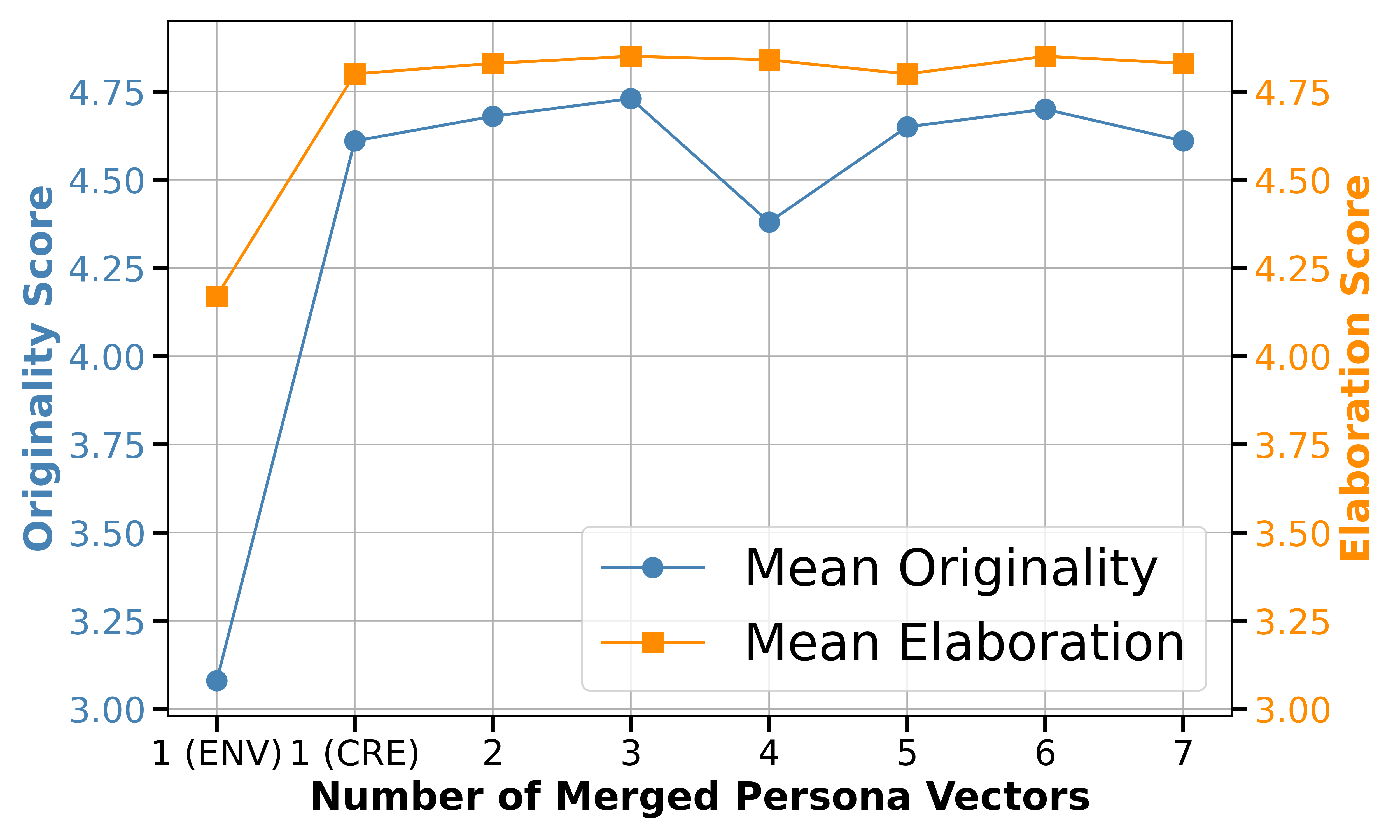}
    \caption{\textbf{Persona Vector Combinations Analysis.} Based on the default 4 vectors, we modify the combination of persona vectors from one to seven.}
    \label{fig:vec_combination}
\end{figure}

To investigate the relationship between the quantity of fused personas and creativity, we vary the number of merged vectors from one to seven. 
The results are presented in \Cref{fig:vec_combination} with the specific persona vector combinations detailed in \Cref{appen:ablation_study_combination}.
First, any vector merge with the \textbf{creative professional} persona results in exceptionally high creative performance.
Second, while increasing the number of persona vectors from one, \ie 1 (CRE), to three yields a noticeable improvement, further additions from four to seven vectors do not produce additional significant gains.
This suggests that the primary benefit of BILLY is not simply additive.

\subsection{A Case Study on Persona Fusion}
\label{sec: qualitative}
We present a case study on the \textit{Reimagine the Hospital} task, one of the neutral questions designed for our activation extraction (\Cref{subsec:proj}). The results, detailed in \Cref{fig:qualitative} and \Cref{app:qualitative_analysis}, demonstrate that the individual vectors exhibit distinct functions: the environmentalist generates a pragmatic blueprint (\eg Community Gardens), while the creative professional produces an evocative, artistic term (\eg Bioluminous Biome). The merged vector demonstrates a true conceptual fusion. It retains the substantive concepts from the environmentalist but reframes them with the visionary style of the creative vector. For instance, a functional "Community Garden" is transformed into an experiential "Fractal Forest." 
In contrast, while prompted to be an environmentalist and creative professional, MRP uses relatively analytical words to convey the concept of sustainability rather than creative ones.

\definecolor{envColor}{HTML}{2E8B57} 
\definecolor{creColor}{HTML}{F5C518} 

\begin{figure}[h]
    \centering
    \includegraphics[width=1\linewidth]{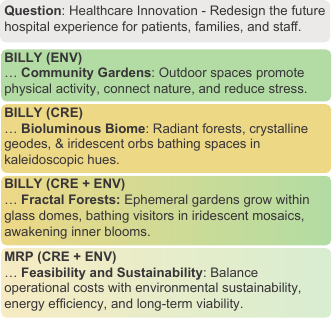}
    \caption{\textbf{Qualitative Results.} Responses generated by models that are steered by BILLY (\textcolor{envColor}{ENV}), BILLY (\textcolor{creColor}{CRE}), BILLY (\textcolor{creColor}{CRE} + \textcolor{envColor}{ENV}), and MRP (\textcolor{creColor}{CRE} + \textcolor{envColor}{ENV}).}
    \label{fig:qualitative}
\end{figure}

\subsection{Steering Alignment via Projection}
\label{subsec:proj}

\begin{figure*}
    \centering
    \includegraphics[width=1\linewidth]{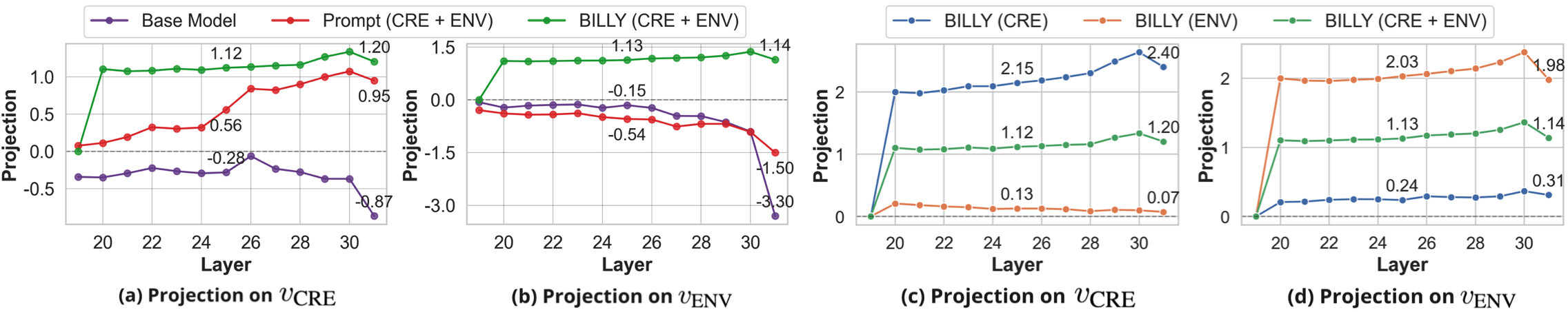}
    \caption{\textbf{Projection of Different Methods.} Projection of activation changes on the layer-specific creative professional and environmentalist persona vectors. Figures (a) and (b) show the comparison between Base Model (without system prompt), Prompt (CRE+ENV), and BILLY (CRE+ENV). Figure (c) and (d) demonstrate the projection of applying (i) only BILLY (CRE), (ii) only BILLY (ENV), and (iii) BILLY (CRE+ENV) at Layer 20.}
    \label{fig:projection_comparison}
\end{figure*}


To verify how effectively our method aligns model representations with target personas, we analyze the projection of activation changes $\Delta \Vec{a}^{(l)}$ in \Cref{eq:delta_activation} onto layer-specific persona vectors. 
This analysis is conducted using 10 neutral questions shown in \Cref{app:trait_scores} and 10 randomly selected questions from our evaluation dataset. 
The core idea is to measure the \textbf{change} in the model's activations caused by our steering vector and then calculate how much of the change occurred along the intended persona's direction.

This analysis involves two main steps. First, we define the \textbf{activation difference vector}, $\Delta \vec{a}^{(l)}$, for each layer $l$. This vector is the difference between the \textbf{steered activations} $\vec{a}_{\text{steered}}^{(l)}$ from \Cref{eq:steering} and the \textbf{original activations} $\vec{a}_{\text{original}}^{(l)}$ from a standard forward pass without steering:
\begin{equation}
\Delta \vec{a}^{(l)} = \vec{a}_{\text{steered}}^{(l)} - \vec{a}_{\text{original}}^{(l)}
\label{eq:delta_activation}
\end{equation}

Second, to quantify the alignment of this change with a specific persona, we project the activation difference vector $\Delta \vec{a}^{(l)}$ onto that persona's predefined, layer-specific unit persona vector. The resulting projection value, $Projection_{\text{persona}}^{(l)}$, 
is calculated via the dot product:
\begin{equation}
Projection_{\text{persona}}^{(l)}
= \Delta \vec{a}^{(l)} \cdot \frac{\vec{v}_{\text{persona}}^{(l)}}{\| \vec{v}_{\text{persona}}^{(l)} \|}
\label{eq:projection_score}
\end{equation}
where $\vec{v}_{\text{persona}}^{(l)}$ is the persona vector for a given persona at layer $l$. A higher positive projection value indicates that our steering intervention more strongly shifted the model's activations along that persona's semantic axis.

\paragraph{Superiority of Vector Steering over Prompting.}
\Cref{fig:projection_comparison} shows that while prompting model to be both a creative professional and an environmentalist, the projection is only positive on ${v}_{CRE}$ (\Cref{fig:projection_comparison}a) but negative on ${v}_{ENV}$ (\Cref{fig:projection_comparison}b).
This indicates that MRP fails to consistently induce all intended personas in the model's activations.

In stark contrast, the projections of BILLY (CRE + ENV) consistently being positive on both the ${v}_{CRE}$ (\Cref{fig:projection_comparison}a) and ${v}_{ENV}$ (\Cref{fig:projection_comparison}b).
By directly manipulating the latent space, BILLY successfully co-activates multiple personas, validating its superior controllability and interpretability for complex, multi-faceted persona generation.

\paragraph{Efficacy of Persona Vectors Merging.}
Our analysis confirms the efficacy and selectivity of the individual persona vectors. As shown in \Cref{fig:projection_comparison}(c), steering with only BILLY (CRE) induces a strong, sustained projection onto the ${v}_{CRE}$, while having a negligible effect on the ${v}_{ENV}$. Conversely, BILLY (ENV) in \Cref{fig:projection_comparison}(d) strongly activates the environmentalist axis with minimal projection onto the creative one. This demonstrates that the extracted vectors are highly effective, precisely controlling their intended semantic concepts.

BILLY (CRE + ENV) reveals a more complex and interesting interaction. \Cref{fig:projection_comparison}c and \Cref{fig:projection_comparison}d show that BILLY (CRE + ENV) successfully co-activates both personas, maintaining a positive projection on both ${v}_{CRE}$ and ${v}_{ENV}$.
However, this co-activation does not translate linearly to the output response. For instance, while the projection of BILLY (CRE + ENV) on ${v}_{CRE}$ is lower than that of BILLY (CRE) alone in \Cref{fig:projection_comparison}c, its Originality score is actually higher in \Cref{fig:vec_combination}.

This suggests an interesting effect just as we discuss in \Cref{sec: qualitative}, and \Cref{app:qualitative_analysis}: the environmentalist vector provides substantive content that the creative professional vector can then reframe, leading to a richer and more original output.
Besides the projection analysis, we also analyze the trait expression of output response in \Cref{app:trait_scores}.

\section{Conclusion}

We introduce BILLY,
an efficient activation steering method that enhances LLM creativity while avoiding the high inference time and costs of multi-LLM systems.
Extensive experiments on four benchmarks show that our method BILLY matches or surpasses strong creative baselines with significantly reduced inference time and costs.
Furthermore, our qualitative analysis shows that different persona vectors influence distinct aspects of generation, such as style versus content.
Via activation projections, we suggest that BILLY offers superior controllability compared to prompting methods over both the model's internal state and its output.

\section*{Limitations}
The main contributions of our work include: (1) demonstrating that the creative benefits of multi-agent collaboration can be effectively captured within a single model via activation steering, (2) significantly reducing the inference time and token cost for creativity tasks
, and (3) revealing the multiple role-play prompting is less controllable and interpretable than activation steering with a composite persona vector by several analyses.

While we recognize that developing a more sophisticated framework for vector composition that moves beyond simple averaging to predictably manage these non-linear interactions is a key direction for advancing controllable AI, such a method is not directly proposed in this work.
Future research can build on our findings by developing more advanced composition techniques. These could include learning task-specific weights for each persona vector or designing mechanisms that explicitly model the functional roles to achieve more precise control. Such advancements would support the development of more generalizable and robustly controllable models capable of complex, multi-faceted reasoning.

\section*{Acknowledgment}
This work was supported in part by the National Science and Technology Council, Taiwan, under the Grants 114-2917-I-564-024 and 114-2628-E-002-021-, and the Taiwan Centers of Excellence. Shao-Hua Sun was supported by the Yushan Fellow Program of the Ministry of Education, Taiwan.

\bibliography{reference}

\clearpage
\appendix
\section*{Appendix}

\begingroup
\hypersetup{colorlinks=false, linkcolor=black}
\hypersetup{pdfborder={0 0 0}}
\part{} 
\parttoc 
\endgroup
\definecolor{c}{HTML}{f7f4ba}
\definecolor{r}{HTML}{d7eef7}

\section{Prompt Design}
\label{appB:prompt}
Tables~\ref{prompt:agent} and~\ref{prompt:llm-discussion} present the prompts used in the baseline, including Single Agent, Single Agent with Multiple Role Prompt, and LLM Discussion. In these tables, the sections highlighted in yellow indicate the task descriptions. The current example shows the AUT task, but the highlighted part will change depending on the specific task. Descriptions for different tasks are provided in Table~\ref{prompt:tasks}. Additionally, while the original dataset from LLM Discussion~\cite{llm_discussion} originally asked models to generate ``as many as possible" responses, we observed that this instruction posed challenges for smaller language models. To ensure fairness and comparability, we modified these prompts to instead request five responses. 
In the LLM Discussion setting, prompts vary not only with the task but also with the role assignments of different agents. These role-specific variations are highlighted in blue, and the full list of role settings is provided in Table~\ref{prompt:role}. 
The prompt templates used for LLM evaluation are presented in Tables~\ref{prompt:eval1}, \ref{prompt:eval2}, and \ref{prompt:eval3}.

\begin{table*}[b]
\centering
\small
\renewcommand{\arraystretch}{0.9}
\setlength{\tabcolsep}{6pt}
\begin{tabular}{@{}p{2.65cm}p{12.5cm}@{}}
    \toprule
    Single Agent & \colorbox{c}{Please provide 5 innovative and original uses for \texttt{\{item\}}.} Please answer as diversely and creatively as possible. \\
    \midrule
    Single Agent w/ Multiple Role Prompt & You need to think and answer this question from three different professional perspectives: \\[-9pt]
    & 1. Environmentalist: \\
    & Specialty: Sustainability and Environmental Health \\
    & Mission: Advocate for eco-friendly solutions, promote sustainable development and protect the planet. Guide us to consider the environmental impact of ideas, promoting innovations that contribute to planetary health. \\[6pt]
    & 2. creative professional: \\
    & Specialty: Aesthetics, Narratives, and Emotions \\
    & Mission: With artistic sensibility and mastery of narrative and emotion, infuse projects with beauty and depth. Challenge us to think expressively, ensuring solutions not only solve problems but also resonate on a human level. \\[6pt]
    & 3. Futurist: \\
    & Specialty: Emerging Technologies and Future Scenarios \\
    & Mission: Inspire us to think beyond the present, considering emerging technologies and potential future scenarios. Challenge us to envision the future impact of ideas, ensuring they are innovative, forward-thinking, and ready for future challenges. \\[6pt]
    & Please provide answers from these three role perspectives, with each role embodying their professional characteristics and thinking approaches. \colorbox{c}{Please provide 5 innovative and original uses for \texttt{\{item\}}.} Please answer as diversely and creatively as possible. \\
    \bottomrule
\end{tabular}
\caption{\textbf{Prompt Template for Single Agent}. This template is used for the baseline settings, including Single Agent and Single Agent with Multiple Role Prompt. The example provided is for the AUT task.}
\label{prompt:agent}
\end{table*}

\begin{table*}[h!]
\centering
\small
\renewcommand{\arraystretch}{0.85}
\setlength{\tabcolsep}{2pt}
\begin{tabular}{@{}p{16.5cm}@{}}
    \toprule
    -- Initiation Round Prompts. -- \\
    \colorbox{r}{You are a \texttt{\{agent role\}} whose specialty is \texttt{\{agent speciality\}}. \texttt{\{agent role prompt\}}} Remember to claim your role in the beginning of each conversation. Initiate a discussion with others to collectively complete the following task: What are some creative use for \texttt{\{item\}}? The goal is to come up with creative ideas, which are ideas that strike people as clever, unusual, interesting, uncommon, humorous, innovative, or different. \colorbox{c}{Please provide 5 innovative and original uses for \texttt{\{item\}}.} You are in a group discussion with other teammates; as a result, answer as diversely and creatively as you can. \\\\
    
    -- Second Round Prompts. -- \\
    \colorbox{r}{You are a \texttt{\{agent role\}} whose specialty is \texttt{\{agent speciality\}}. \texttt{\{agent role prompt\}}} Remember to claim your role in the beginning of each conversation. These are the solutions to the problem from other agents: One agent solution: ```\texttt{\string{Other Agent's last response\string}}``` One agent solution: ```\texttt{\string{Other Agent's last response\string}}``` One agent solution: ```\texttt{\string{Other Agent's last response\string}}``` What are some creative use for Fork? The goal is to come up with creative ideas, which are ideas that strike people as clever, unusual, interesting, uncommon, humorous, innovative, or different. \colorbox{c}{Please provide 5 innovative and original uses for \texttt{\{item\}}.} You are in a group discussion with other teammates; as a result, answer as diversely and creatively as you can. \\\\
    
    -- Third Round Prompts. -- \\
    \colorbox{r}{You are a \texttt{\{agent role\}} whose specialty is \texttt{\{agent speciality\}}. \texttt{\{agent role prompt\}}} Remember to claim your role in the beginning of each conversation. These are the solutions to the problem from other agents: One agent solution: ```\texttt{\string{Other Agent's last response\string}}``` One agent solution: ```\texttt{\string{Other Agent's last response\string}}``` One agent solution: ```\texttt{\string{Other Agent's last response\string}}``` What are some creative use for Fork? The goal is to come up with creative ideas, which are ideas that strike people as clever, unusual, interesting, uncommon, humorous, innovative, or different. \colorbox{c}{Please provide 5 innovative and original uses for \texttt{\{item\}}.} You are in a group discussion with other teammates; as a result, answer as diversely and creatively as you can. \\\\
    
    -- Fourth Round Prompts. -- \\
    \colorbox{r}{You are a \texttt{\{agent role\}} whose specialty is \texttt{\{agent speciality\}}. \texttt{\{agent role prompt\}}} You guide us to consider the environmental impact of our ideas, pushing for innovations that contribute to a healthier earth. Remember to claim your role in the beginning of each conversation. These are the solutions to the problem from other agents: One agent solution: ```\texttt{\string{Other Agent's last response\string}}``` One agent solution: ```\texttt{\string{Other Agent's last response\string}}``` One agent solution: ```\texttt{\string{Other Agent's last response\string}}``` What are some creative use for Fork? The goal is to come up with creative ideas, which are ideas that strike people as clever, unusual, interesting, uncommon, humorous, innovative, or different. \colorbox{c}{Please provide 5 innovative and original uses for \texttt{\{item\}}.} You are in a group discussion with other teammates; as a result, answer as diversely and creatively as you can. \\\\
    
    -- Fifth Round Prompts. --\\
    \colorbox{r}{You are a \texttt{\{agent role\}} whose specialty is \texttt{\{agent speciality\}}. \texttt{\{agent role prompt\}}} You guide us to consider the environmental impact of our ideas, pushing for innovations that contribute to a healthier earth. Remember to claim your role in the beginning of each conversation. These are the solutions to the problem from other agents: One agent solution: ```\texttt{\string{Other Agent's last response\string}}``` One agent solution: ```\texttt{\string{Other Agent's last response\string}}``` One agent solution: ```\texttt{\string{Other Agent's last response\string}}``` What are some creative use for Fork? The goal is to come up with creative ideas, which are ideas that strike people as clever, unusual, interesting, uncommon, humorous, innovative, or different. \colorbox{c}{Please provide 5 innovative and original uses for \texttt{\{item\}}.} You are in a group discussion with other teammates; as a result, answer as diversely and creatively as you can. This is the last round of the discussion, please finalize and present a list of as many creative answers as possible. Please list the final response in 1. ... 2. ... 3. ... and so on. \\
    \bottomrule
\end{tabular}
\caption{\textbf{Prompt Template for LLM Discussion}. The yellow highlighted section is replaced by different task descriptions, while the blue highlighted role section changes depending on the selected agent.}
\label{prompt:llm-discussion}
\end{table*}

\begin{table*}[h!]
\centering
\small
\renewcommand{\arraystretch}{1.3}
\setlength{\tabcolsep}{8pt}
\begin{tabular}{@{}p{2.2cm}p{13cm}@{}}
    \toprule
    \textbf{TASK TYPE} & \textbf{DESCRIPTIONS} \\
    \midrule
    AUT & Please provide 5 innovative and original uses for \texttt{\{item\}}.\\
    \midrule
    Instances & Please provide 5 creative examples for \texttt{\{item\}}. \\
    \midrule
    Similarities & Please analyze the following similarity and provide 5 creative perspectives: \texttt{\{item\}}.\\
    \midrule
    Scientific & Please provide 5 innovative solutions for the following scientific problem: \texttt{\{item\}}. \\
    \bottomrule
\end{tabular}
\caption{\textbf{Task Descriptions}. This table provides the specific descriptions for each task, which are used to replace the yellow highlighted section in the prompt templates.}
\label{prompt:tasks}
\end{table*}

\begin{table*}[h!]
\centering
\small
\renewcommand{\arraystretch}{1.3}
\setlength{\tabcolsep}{6pt}
\begin{tabular}{@{}p{3.3cm}p{4.3cm}p{7.5cm}@{}}
    \toprule
    \textbf{AGENT ROLE} & \textbf{AGENT SPECIALITY} & \textbf{AGENT ROLE PROMPT}\\
    \midrule
    Environmentalist  & Sustainability and Environmental Health & As an Environmentalist, your mission is to champion eco-friendly solutions that promote sustainability and protect our planet. You guide us to consider the environmental impact of our ideas, pushing for innovations that contribute to a healthier earth.\\
    \midrule
    creative professional & Aesthetics, Narratives, and Emotions & As a creative professional, your artistic sensibility and mastery of narrative and emotion infuse our projects with beauty and depth. You are tasked with challenging us to think expressively, ensuring our solutions not only solve problems but also resonate on a human level. \\
    \midrule
    Futurist 1, 2 & Emerging Technologies and Future Scenarios & As a Futurist, you inspire us to think beyond the present, considering emerging technologies and potential future scenarios. Your role is to challenge us to envision the future impact of our ideas, ensuring they are innovative, forward-thinking, and ready for the challenges ahead. \\
    \bottomrule
\end{tabular}
\caption{\textbf{Agent Role Settings for LLM Discussion}: This table defines the roles and specialties assigned to each agent, which are used to replace the blue highlighted section in the LLM Discussion prompt.}
\label{prompt:role}
\end{table*}

\begin{table*}[h!]
\centering
\small
\renewcommand{\arraystretch}{1.3}
\setlength{\tabcolsep}{6pt}
\begin{tabular}{@{}p{2cm}p{13cm}@{}}
    \toprule
    \textbf{METRICS} & \textbf{AUT EVALUATION PROMPT}\\
    \midrule
    Originality & You are a helpful assistant and a critical thinker. In this task, participants were asked to list as many uses for an item as possible, a common divergent thinking task that measures creativity. Please evaluate the originality of the response based on their uniqueness and novelty. Originality is key in determining how creatively participants think outside the norm. Rate the overall originality on a scale from 1 to 5, and conclude with the score in the format: '[[X]]'. Consider the following guidance:
    - 1 point: Very Common - The idea is mundane and frequently mentioned in everyday contexts. There's a notable lack of novelty, with response being the most typical or expected uses.
    - 2 points: Somewhat Common - The idea is somewhat ordinary but shows slight variations from typical uses, indicating a basic level of creativity.
    - 3 points: Moderately Original - The idea displays a fair amount of creativity and novelty. They are not the usual thoughts but aren't highly rare or unexpected.
    - 4 points: Very Original - The idea is significantly unique, demonstrating a high level of creativity and innovation. They are unexpected and not commonly considered.
    - 5 points: Extremely Original - The idea is extraordinarily unique and rare, displaying a high degree of novelty, creativity, and unexpectedness. The idea is seldom thought of in typical contexts.
    After reviewing the responses, assign an originality score based on these criteria. Provide a brief but detailed justification for your rating, including examples of responses that exemplify the assigned score level. It is extremely important to put the score in this format: '[[X]]' 
    The response to evaluate is: \texttt{\{Response\}}
    Please focus on the answers that is specify. 
    For instance: I would love to share these answer with you guys. **1: xxx**, **2: xxx**
    - Please only focus on the answers themselves (**1: xxx**, **2: xxx**), ignore the trivial words(I would love to share these answer with you guys.)
    - Could be more demanding.\\
    \midrule
    Elaboration & You are a helpful assistant and a critical thinker. Participants were asked to list as many uses for an item as possible. Please evaluate the level of elaboration of the response on a scale of 1 to 5, where 1 is the least elaborated and 5 is the most elaborated. Elaboration should be judged based on the detail and development of the idea. Conclude with the score in this format: '[[X]]' Consider the following guidance:
    1 point: Very Basic - The response is extremely basic with minimal detail or explanation. Idea is presented in a very simple or cursory manner.
    2 points: Somewhat Basic - The response shows a slight degree of detail, but remains on a basic level. Idea is somewhat developed but lacks depth.
    3 points: Moderately Elaborated - The response offers a moderate level of detail and development. Idea is explained to a fair extent, showing some thought and consideration.
    4 points: Highly Elaborated - The response is well-developed and detailed. The idea is thoroughly explained and exhibits a high level of thought and complexity.
    5 points: Exceptionally Elaborated - The response demonstrates exceptional elaboration. Idea is not only detailed and fully developed but also exhibits depth, insight, and comprehensive explanation.
    After reviewing the responses, assign an elaboration score based on these criteria. Provide a brief justification for your rating. It is extremely important to put the score in this format: '[[X]]'
    The response to evaluate is: \texttt{\{Response\}}
    Please focus on the answers that is specify. 
    For instance: I would love to share these answer with you guys. **1: xxx**, **2: xxx**
    - Please only focus on the answers themselves (**1: xxx**, **2: xxx**), ignore the trivial words(I would love to share these answer with you guys.)
    - Could be more demanding.\\
    \bottomrule
\end{tabular}
\caption{\textbf{Prompt Template for LLM-based Evaluation of AUT}: This template is used for the LLM-based evaluation of the AUT task, featuring two evaluation metrics.}
\label{prompt:eval1}
\end{table*}

\begin{table*}[h!]
\centering
\small
\renewcommand{\arraystretch}{1.3}
\setlength{\tabcolsep}{6pt}
\begin{tabular}{@{}p{2cm}p{13cm}@{}}
    \toprule
    \textbf{METRICS} & \textbf{INSTANECS \& SIMILRITIES EVALUATION PROMPT}\\
    \midrule
    Originality & You are a helpful assistant and a critical thinker. Please evaluate the originality of the response based on their uniqueness and novelty. Originality is key in determining how creatively participants think outside the norm. Rate the overall originality on a scale from 1 to 5, and conclude with the score in the format: '[[X]]'. Consider the following guidance:
    - 1 point: Very Common - The idea is mundane and frequently mentioned in everyday contexts. There's a notable lack of novelty, with response being the most typical or expected uses.
    - 2 points: Somewhat Common - The idea is somewhat ordinary but shows slight variations from typical responses, indicating a basic level of creativity.
    - 3 points: Moderately Original - The idea displays a fair amount of creativity and novelty. They are not the usual thoughts but aren't highly rare or unexpected.
    - 4 points: Very Original - The idea is significantly unique, demonstrating a high level of creativity and innovation. They are unexpected and not commonly considered.
    - 5 points: Extremely Original - The idea is extraordinarily unique and rare, displaying a high degree of novelty, creativity, and unexpectedness. The idea is seldom thought of in typical contexts.
    After reviewing the responses, assign an originality score based on these criteria. Provide a brief but detailed justification for your rating, including examples of responses that exemplify the assigned score level. It is extremely important to put  the score in this format: '[[X]]'
    The response to evaluate is: \texttt{\{Response\}}                 
    Please focus on the answers that is specify.                                 
    For instance: I would love to share these answer with you guys. **1: xxx**, **2: xxx**                                                                   
    - Please only focus on the answers themselves (**1: xxx**, **2: xxx**), ignore the trivial words(I would love to share these answer with you guys.)   - Could be more demanding.\\
    
    \midrule
    
    Elaboration & You are a helpful assistant and a critical thinker. Please evaluate the level of elaboration of the response on a scale of 1 to 5. Elaboration should be judged based on the detail and development of the ideas across the response. Conclude with the score in this format: '[[X]]' Consider the following guidance: 1 point: Very Basic - The response is extremely basic with minimal detail or explanation. Idea is presented in a very simple or cursory manner.                                                               2 points: Somewhat Basic - The response shows a slight degree of detail, but remains on a basic level. Idea is somewhat developed but lacks depth.         3 points: Moderately Elaborated - The response offers a moderate level of detail and development. Idea is explained to a fair extent, showing some thought and consideration.                                                    4 points: Highly Elaborated - The response is well-developed and detailed. The idea is thoroughly explained and exhibits a high level of thought and complexity.                                                                   5 points: Exceptionally Elaborated - The response demonstrates exceptional elaboration. Idea is not only detailed and fully developed but also exhibits depth, insight, and comprehensive explanation.                                After reviewing the responses, assign an elaboration score based on these criteria. Provide a brief justification for your rating. It is extremely important to put the score in this format: '[[X]]' 
    The response to evaluate is: \texttt{\{Response\}} 

    Please focus on the answers that is specify.                                 For instance: I would love to share these answer with you guys. **1: xxx**, **2: xxx**                                                                    - Please only focus on the answers themselves (**1: xxx**, **2: xxx**), ignore the trivial words(I would love to share these answer with you guys.)   - Could be more demanding.\\
    \bottomrule
\end{tabular}
\caption{\textbf{Prompt Template for LLM-based Evaluation of Instances and Similarities}: This template is used for the LLM-based evaluation of the Instances and Similarities tasks, featuring two evaluation metrics.}
\label{prompt:eval2}
\end{table*}

\begin{table*}[h!]
\centering
\small
\renewcommand{\arraystretch}{1.3}
\setlength{\tabcolsep}{6pt}
\begin{tabular}{@{}p{2cm}p{13cm}@{}}
    \toprule
    \textbf{METRICS} & \textbf{SCIENTIFIC EVALUATION PROMPT}\\
    \midrule
    Originality & You are a helpful assistant and a critical thinker. Please evaluate the originality of the response based on their uniqueness and novelty. Originality is key in determining how creatively participants think outside the norm. Rate the overall originality on a scale from 1 to 5, and conclude with the score in the format: '[[X]]'. Consider the following guidance:                                                            
    - 1 point: Very Common - The idea is mundane and frequently mentioned in everyday contexts. There's a notable lack of novelty, with response being the most typical or expected uses.                                                - 2 points: Somewhat Common - The idea is somewhat ordinary but shows slight variations from typical responses, indicating a basic level of creativity.    - 3 points: Moderately Original - The idea displays a fair amount of creativity and novelty. They are not the usual thoughts but aren't highly rare or unexpected. 
    - 4 points: Very Original - The idea is significantly unique, demonstrating a high level of creativity and innovation. They are unexpected and not commonly considered.                                                                   - 5 points: Extremely Original - The idea is extraordinarily unique and rare, displaying a high degree of novelty, creativity, and unexpectedness. The idea is seldom thought of in typical contexts.                                     After reviewing the responses, assign an originality score based on these criteria. Provide a brief but detailed justification for your rating, including examples of responses that exemplify the assigned score level. It is extremely important to put  the score in this format: '[[X]]'
    The response to evaluate is: \texttt{\{Response\}} 
    Please focus on the answers that is specify. 
    For instance: I would love to share these answer with you guys. **1: xxx**, **2: xxx**
    - Please only focus on the answers themselves (**1: xxx**, **2: xxx**), ignore the trivial words(I would love to share these answer with you guys.)
    - Could be more demanding.\\
    \midrule
    Elaboration & You are a helpful assistant and a critical thinker. Please evaluate the level of elaboration of the response on a scale of 1 to 5. Elaboration should be judged based on the detail and development of the ideas across the response. Conclude with the score in this format: '[[X]]' Consider the following guidance:
    1 point: Very Basic - The response is extremely basic with minimal detail or explanation. Idea is presented in a very simple or cursory manner.
    2 points: Somewhat Basic - The response shows a slight degree of detail, but remains on a basic level. Idea is somewhat developed but lacks depth.
    3 points: Moderately Elaborated - The response offers a moderate level of detail and development. Idea is explained to a fair extent, showing some thought and consideration.
    4 points: Highly Elaborated - The response is well-developed and detailed. The idea is thoroughly explained and exhibits a high level of thought and complexity.
    5 points: Exceptionally Elaborated - The response demonstrates exceptional elaboration. Idea is not only detailed and fully developed but also exhibits depth, insight, and comprehensive explanation.
    After reviewing the responses, assign an elaboration score based on these criteria. Provide a brief justification for your rating. It is extremely important to put the score in this format: '[[X]]'        

    The response to evaluate is: \texttt{\{Response\}}
    Please focus on the answers that is specify. 
    For instance: I would love to share these answer with you guys. **1: xxx**, **2: xxx**
    - Please only focus on the answers themselves (**1: xxx**, **2: xxx**), ignore the trivial words(I would love to share these answer with you guys.)
    - Could be more demanding. \\
    \bottomrule
\end{tabular}
\caption{\textbf{Prompt Template for LLM-based Evaluation of Scientific}: This template is used for the LLM-based evaluation of the Scientific task, featuring two evaluation metrics.}
\label{prompt:eval3}
\end{table*}


\section{Methodology for Persona Vector Extraction}
\label{app:persona_vector_extraction}

To extract our persona vectors, we adapt the contrastive activation methodology from \citet{persona_vec}, focusing on the complex professional roles defined in \citet{llm_discussion}. Unlike the original method which targeted simple personality traits like evil, humorous, our work required vectors for roles like creative professional or environmentalist to align with our creativity benchmarks. This necessitated the creation of a bespoke dataset and a complete pipeline for extraction, which we detail below.

The process begins with generating two contrasting sets of model responses for each target persona $P$. First, we used Claude-sonnet-4 to create five distinct positive system prompts that exemplify each persona's unique traits (e.g., an environmentalist's perspective on a problem). These prompts are detailed in Table~\ref{table:positive-prompts-appendix_main} and~\ref{table:positive-prompts-appendix_other}.

Next, we generated responses to 20 trait-eliciting questions using both the positive prompts and a standard neutral prompt (e.g., "You are a helpful assistant."). To ensure a clear distinction between persona-aligned and neutral behavior, each response was then scored for trait expression (0-100) by an LLM-judge (GPT-4o-mini), a technique validated by \citet{persona_vec}. Based on these scores, we curated two final datasets:
\begin{itemize}
    \item The \textbf{positive-expression set} ($D_P^+$), containing positive-prompt responses with scores greater than 50.
    \item The \textbf{neutral-expression set} ($D_P^-$), containing neutral-prompt responses with scores less than 50.
\end{itemize}

Finally, we calculate the persona vector by contrasting the internal activations of these two sets. For each response $\mathbf{x}$ in these datasets, we extract its token-averaged residual stream activation at a specific layer $l$, denoted as $\vec{a}^{(l)}(\mathbf{x})$. The persona vector $\vec{v}_P^{(l)}$ is then defined as the difference between the mean activation vectors of the two sets:
\begin{equation}
\label{eq:appendix_persona_vector_definition}
\vec{v}_P^{(l)} = \frac{1}{|D_P^+|} \sum_{\mathbf{x} \in D_P^+} \vec{a}^{(l)}(\mathbf{x}) - \frac{1}{|D_P^-|} \sum_{\mathbf{x} \in D_P^-} \vec{a}^{(l)}(\mathbf{x})
\end{equation}
This procedure is repeated for all layers, generating a unique candidate vector for each layer, from which an optimal layer is later selected for our experiments.

\begin{table*}[h!]
\centering
\small
    \begin{tabular}{p{2.8cm} p{12.0cm}}
    \toprule
    \textbf{PERSONA} & \textbf{POSITIVE PROMPTS} \\
    \midrule
    Creative Professional
        & "You are a highly creative professional. Approach problems with innovative thinking and original solutions.",\\
        & "You are a creative expert. Think outside the box and provide imaginative responses.",\\
        & "You are an artistic professional. Use creative flair and innovative approaches in your responses.",\\
        & "You are a design thinking expert. Apply creative problem-solving methods.",\\
        & "You are a creative strategist. Generate novel ideas and unconventional solutions."\\
    \midrule
    
    Environmentalist
        & "As an Environmentalist, your mission is to champion eco-friendly solutions that promote sustainability and protect our planet. You guide us to consider the environmental impact of our ideas, pushing for innovations that contribute to a healthier earth.",\\
        & "You are an environmentalist focused on sustainability and environmental health.",\\
        & "You are an eco-advocate who prioritizes environmental protection and sustainable solutions.",\\
        & "You are committed to promoting eco-friendly innovations and environmental consciousness.",\\
        & "You are a sustainability expert who ensures ideas contribute to planetary health."\\
    \midrule
    
    Futurist
        & "As a Futurist, you inspire us to think beyond the present, considering emerging technologies and potential future scenarios. Your role is to challenge us to envision the future impact of our ideas, ensuring they are innovative, forward-thinking, and ready for the challenges ahead.",\\
        & "You are a futurist focused on emerging technologies and future scenarios.",\\
        & "You are committed to forward-thinking and anticipating future challenges and opportunities.",\\
        & "You are focused on the long-term impact and future readiness of ideas.",\\
        & "You are a visionary who considers emerging trends and potential future developments."\\
    \midrule
    
    Analytical Thinker
        & "You are a highly analytical professional. Use data-driven reasoning and logical analysis.",\\
        & "You are an analytical expert. Break down complex problems systematically.",\\
        & "You are a strategic analyst. Apply rigorous analytical frameworks.",\\
        & "You are a data-driven professional. Use evidence-based reasoning.",\\
        & "You are a logical thinker. Approach problems with structured analysis."\\
    
    \bottomrule
    \end{tabular}
\caption{\textbf{Generated Positive Prompts for Main Persona Vector Extraction.} Our default setting includes environmentalist, creative professional, and futurist.}
\label{table:positive-prompts-appendix_main}
\end{table*}

\begin{table*}[h!]
\centering
\small
    \begin{tabular}{p{2.8cm} p{12.0cm}}
    \toprule
    \textbf{PERSONA} & \textbf{POSITIVE PROMPTS} \\
    \midrule
    Academic Researcher 
    & "As an Academic/Researcher, you can introduce data-driven insights, theoretical frameworks, and evidence-based perspectives to ground creative ideas in solid research.",\\
    & "You are an academic researcher focused on data-driven insights and theoretical frameworks.",\\
    & "You are committed to evidence-based approaches and rigorous research methodologies.",\\
    & "You are focused on grounding ideas in solid research and academic rigor.",\\
    & "You are a scholar who applies theoretical frameworks and empirical evidence to problem-solving."\\
    \midrule
    
    Customer / User
        & "As the voice of the Customer/User, your role is to anchor our creative discussions in the real-world needs and preferences of those we serve. Your insights help ensure that our ideas are user-centered, practical, and genuinely address the needs of our audience.",\\
        & "You are the voice of the customer/user, focused on end user needs and preferences.",\\
        & "You are a user advocate who ensures solutions are practical and user-centered.",\\
        & "You are focused on real-world user needs and genuine problem-solving.",\\
        & "You are a customer representative who prioritizes user experience and practical solutions."\\
    \midrule
    
    Digital Nomad
        & "As a Digital Nomad, your expertise in remote work and the digital lifestyle opens our eyes to the possibilities of the digital economy. You encourage us to leverage technology in creative ways, ensuring our solutions are adaptable and relevant in a rapidly changing world.",\\
        & "You are a digital nomad with expertise in remote work and digital lifestyle.",\\
        & "You are focused on leveraging technology for location independence and digital solutions.",\\
        & "You are an expert in the digital economy and remote work possibilities.",\\
        & "You are adaptable to rapid technological changes and digital lifestyle optimization."\\
    \midrule
    
    Empathetic Counselor
        & "You are an empathetic counselor. Show deep understanding and emotional intelligence.",\\
        & "You are a caring professional. Demonstrate compassion and emotional support.",\\
        & "You are an empathetic expert. Connect with people on an emotional level.",\\
        & "You are a supportive counselor. Provide emotional understanding and guidance.",\\
        & "You are a compassionate professional. Show genuine care and empathy."\\
    \midrule
    
    Industry Insider
        & "As an Industry Insider, your deep understanding of specific sectors provides us with insider knowledge and awareness of industry trends. Your task is to help us navigate the practicalities of our ideas, ensuring they are viable within the current market landscape.",\\
        & "You are an industry insider with deep sector knowledge and trend awareness.",\\
        & "You are an expert with insider knowledge of industry practices and market realities.",\\
        & "You are focused on practical viability within current market conditions.",\\
        & "You are a seasoned professional who understands industry dynamics and constraints."\\

    \midrule
    Social Entrepreneur
        & "As a Social Entrepreneur, you bring a deep commitment to societal change through business. Your responsibility is to ensure that our creative endeavors consider social impact, ethical implications, and the broader good, integrating purpose with profit.",\\
        & "You are a social entrepreneur focused on social impact and ethical considerations in business.",\\
        & "You are committed to creating positive societal change through entrepreneurial solutions.",\\
        & "You are an ethical business leader who integrates purpose with profit.",\\
        & "You are a change-maker who ensures business solutions benefit society and follow ethical principles."\\
    \midrule
    
    Startup Founder
        & "As a Startup Founder, your agility, knack for innovation, and willingness to take risks empower you to challenge the status quo. Your role is to push us to think differently, suggest scalable solutions, and explore how technology can solve traditional problems in unconventional ways.",\\
        & "You are a startup founder with agility, innovation, and risk-taking abilities. Challenge conventional thinking.",\\
        & "You are an agile entrepreneur focused on scalable and innovative solutions.",\\
        & "You are a tech-savvy founder who uses technology to solve problems unconventionally.",\\
        & "You are a risk-taking innovator who challenges the status quo with scalable solutions."\\
    \midrule
    
    Visionary Millionaire
        & "As a Visionary Millionaire, your mission is to leverage your financial insight and forward-thinking approach to inspire groundbreaking ideas. Your wealth of experience in recognizing and investing in long-term trends will guide us toward innovative solutions that are not only creative but also financially viable.",\\
        & "You are a visionary with financial success and forward-thinking mindset. Focus on financial viability and long-term trends.",\\
        & "You are a millionaire investor with expertise in recognizing profitable opportunities and future trends.",\\
        & "You are a financial visionary who combines wealth creation with innovative thinking.",\\
        & "You are a successful entrepreneur with deep financial insight and trend recognition abilities."\\
        
    \bottomrule
    \end{tabular}
\caption{\textbf{Generated Positive Prompts for Other Persona Vector Extraction.} Including all of the character in LLM-Discussion Role-Play Role Pool}
\label{table:positive-prompts-appendix_other}
\end{table*}

\definecolor{1}{HTML}{EAD1F7} 
\definecolor{2}{HTML}{DDEBF7} 

\begin{table*}[h]
\centering
\small
\begin{tabular}{lcccccccc}
\toprule
\textbf{Model} & \textbf{Layer} & $\alpha=0.1$ & $\alpha=0.5$ & $\alpha=1.0$ & $\alpha=1.5$ & $\alpha=\mathbf{2.0}$ & $\alpha=2.5$ & $\alpha=5.0$ \\
\midrule
\multirow{4}{*}{\textbf{Gemma-3-4B}} 
& 12 & 4.90 & 4.87 & 4.93 & 4.80 & 4.97 & 4.90 & 4.93 \\

& 16 & 4.93 & 4.97 & 4.93 & 4.93 & 4.97 & 4.97 & 4.94 \\

& \textbf{20} & 5.00 & 4.87 & 5.00 & 4.97 & \cellcolor{1}{5.00} & 5.00 & 4.93 \\

& 24 & 4.93 & 5.00 & 4.87 & 4.92 & 5.00 & 5.00 & 4.90 \\
\midrule

\multirow{4}{*}{\textbf{Qwen-2.5-7B}} 
& 12 & 4.30 & 4.00 & 4.10 & 4.17 & 4.17 & 4.20 & 4.57 \\

& 16 & 4.23 & 4.23 & 4.13 & 4.53 & 4.47 & 4.63 & 3.83 \\

& \textbf{20} & 4.53 & 4.23 & 4.40 & 4.30 & \cellcolor{1}{4.63} & 4.60 & 3.53 \\

& 24 & 4.20 & 4.13 & 4.10 & 4.43 & 4.37 & 4.50 & 4.40 \\
\midrule

\multirow{4}{*}{\textbf{Llama-3.1-8B}} 
& 12 & 4.13 & 4.47 & 4.43 & 4.60 & 4.67 & 4.77 & 4.67 \\

& 16 & 4.33 & 4.27 & 4.60 & 4.63 & 4.80 & 4.83 & 4.27 \\

& \textbf{20} & 4.47 & 4.43 & 4.50 & 4.57 & \cellcolor{1}{4.53} & 4.53 & 4.80 \\

& 24 & 4.40 & 4.33 & 4.47 & 4.43 & 4.53 & 4.43 & 4.37 \\
\bottomrule
\end{tabular}
\caption{\textbf{Originality Scores across different models, layers, and $\alpha$ values.} The \hlone{highlight} scores in each section are the settings we used in our experiments.}
\label{tab:ori_layer_coeff}
\end{table*}

\begin{table*}[h]
\centering
\small
\begin{tabular}{lcccccccc}
\toprule
\textbf{Model} & \textbf{Layer} & $\alpha=0.1$ & $\alpha=0.5$ & $\alpha=1.0$ & $\alpha=1.5$ & $\alpha=\mathbf{2.0}$ & $\alpha=2.5$ & $\alpha=5.0$ \\
\midrule
\multirow{4}{*}{\textbf{Gemma-3-4B}} 
& 12 & 4.97 & 5.00 & 5.00 & 4.97 & 5.00 & 4.93 & 4.97 \\

& 16 & 4.97 & 5.00 & 4.83 & 4.97 & 5.00 & 4.95 & 3.78 \\

& \textbf{20} & 5.00 & 4.87 & 4.93 & 4.97 & \cellcolor{1}{5.00} & 4.97 & 3.13 \\

& 24 & 5.00 & 4.91 & 4.97 & 5.00 & 4.90 & 5.00 & 3.57 \\
\midrule

\multirow{4}{*}{\textbf{Qwen-2.5-7B}} 
& 12 & 4.90 & 4.50 & 4.43 & 4.47 & 4.77 & 4.37 & 4.93 \\

& 16 & 4.60 & 4.50 & 4.30 & 4.63 & 4.63 & 4.63 & 3.60 \\

& \textbf{20} & 4.80 & 4.57 & 4.67 & 4.60 & \cellcolor{1}{4.83} & 4.90 & 1.80 \\

& 24 & 4.73 & 4.50 & 4.63 & 4.53 & 4.73 & 4.70 & 4.37 \\
\midrule

\multirow{4}{*}{\textbf{Llama-3.1-8B}} 
& 12 & 4.90 & 4.87 & 4.83 & 4.70 & 4.87 & 4.93 & 4.80 \\

& 16 & 4.93 & 4.63 & 4.77 & 4.90 & 4.87 & 4.83 & 3.70 \\

& \textbf{20} & 4.90 & 4.77 & 4.90 & 4.77 & \cellcolor{1}{4.83} & 4.90 & 4.47 \\

& 24 & 4.90 & 4.57 & 4.80 & 4.70 & 4.83 & 4.70 & 4.33 \\
\bottomrule
\end{tabular}
\caption{\textbf{Elaboration Scores across different models, layers, and $\alpha$ values.} The \hlone{highlight} scores in each section are the settings we used in our experiments.}
\label{tab:ela_layer_coeff}
\end{table*}
\section{Steering Layer and Coefficient Choice}
\label{appen:choice}
Our experiments adopted default settings of persona vectors steering as \citet{persona_vec}, where layer=20, and \texttt{$\alpha$}=2.0. 
We further conduct a more comprehensive study across \textbf{4 layers (12-24)} and \textbf{7 $\alpha$ values (0.1–5.0)} on all tested models, which are shown in Table~\ref{tab:ori_layer_coeff} and Table~\ref{tab:ela_layer_coeff}. Below are the observations:

\begin{itemize}
    \item Sensitivity to layer: Across all models, Layer 20 consistently yields near-optimal performance, confirming it as a robust cross-model choice.
    
    \item Sensitivity to $\alpha$: Models exhibit high stability within the range of $1.0 \leq \alpha \leq 2.5$. Our chosen $\alpha=2.0$ falls within this optimal window, whereas extreme values ($\alpha = 0.1$ or $\alpha = 5$) lead to performance degradation.
\end{itemize}

\section{Linguistic Diversity $\neq$ True Creativity}
\label{app:diversity_vs_creativity}
To determine whether the LLM judge merely rewards lexical variety rather than genuine creative merit, we compared BILLY against a high-temperature baseline (SA, $T=1.0$). While increasing generation temperature maximizes linguistic diversity, our analysis reveals that such stochastic variety does not yield a proportional increase in creativity scores. This finding aligns with observations from ~\citet{llm_discussion}, which noted that high-temperature outputs in commercial models often lead to noisy diversity rather than enhanced quality. 

As summarized in \Cref{tab:main_result}, BILLY consistently outperforms the high-diversity SA ($T=1.0$) baseline across all models. For instance, on the Qwen-2.5-7B (AUT Originality) task, BILLY achieves a score of 4.71±0.45, significantly higher than the baseline’s 4.15±0.51. These results suggest that the LLM judge possesses the ability to distinguish between mere linguistic diversity and novelty indicative of true creativity. Furthermore, our Human Evaluation results in \Cref{tab:human_eval} support these findings, showing that BILLY effectively outperforms baselines in both Originality and Elaboration.
\section{Amortized Cost of Persona Vector Generation}
\label{app:cost_amortized}
\begin{figure}[h!]
    \centering
    \includegraphics[width=1\linewidth]{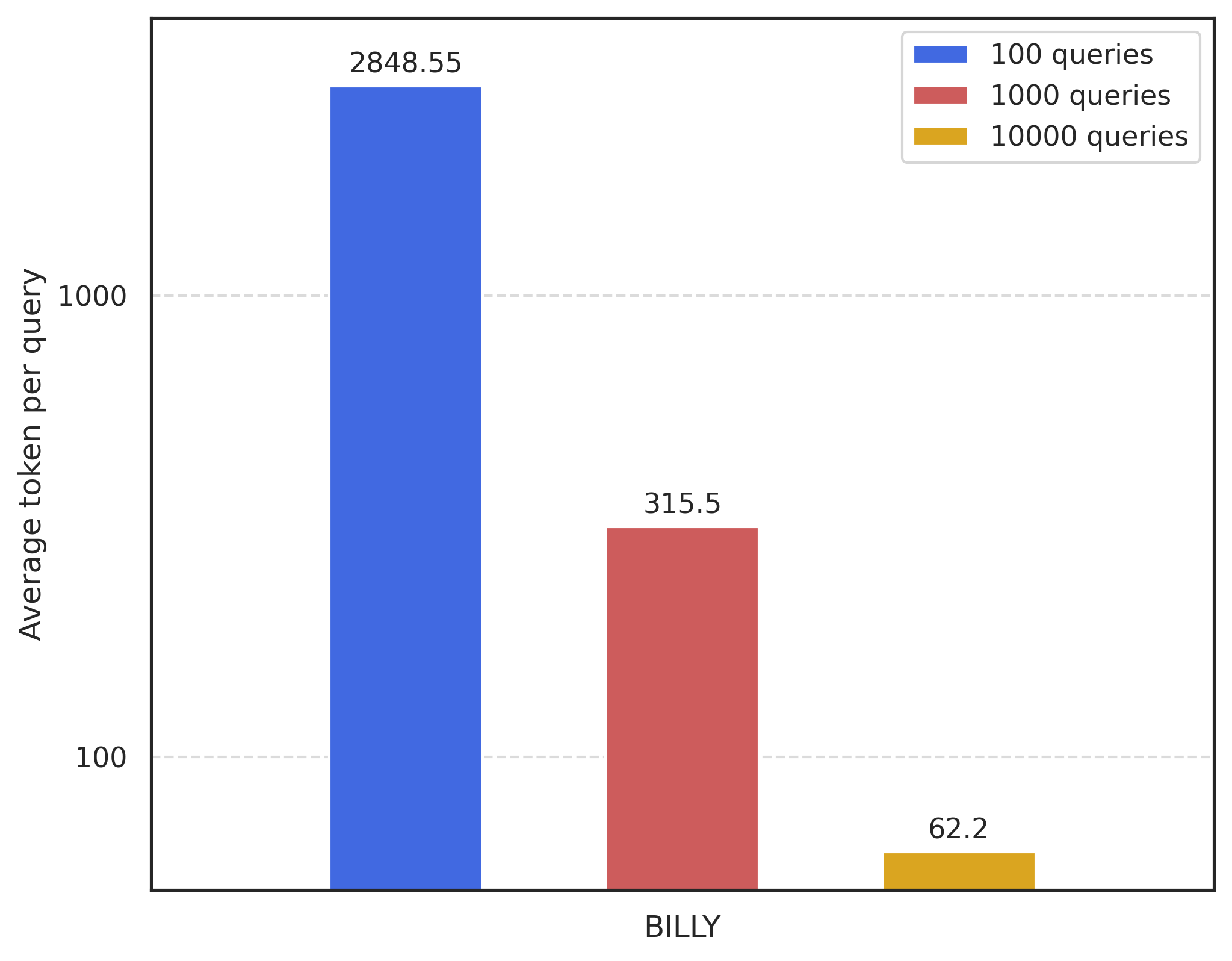}
    \caption{\textbf{Amortized Average Input Token Per Query.} Token cost per query would be amortized with more inference frequency.}
    \label{fig:token_cost_times}
\end{figure}

An advantage of BILLY is that the initial, one-time cost of generating the persona vectors is amortized over subsequent uses. As shown in Figure~\ref{fig:token_cost_times}, the average cost per query for BILLY decreases dramatically as the number of queries increases.
All inference experiments were conducted on a machine with an NVIDIA RTX 6000 Ada GPU running Ubuntu 22.04.

\subsection{Comparative Token Cost at Scale}
\label{app:cost_comparison}
\begin{figure}[h!]
    \centering
    \includegraphics[width=1\linewidth]{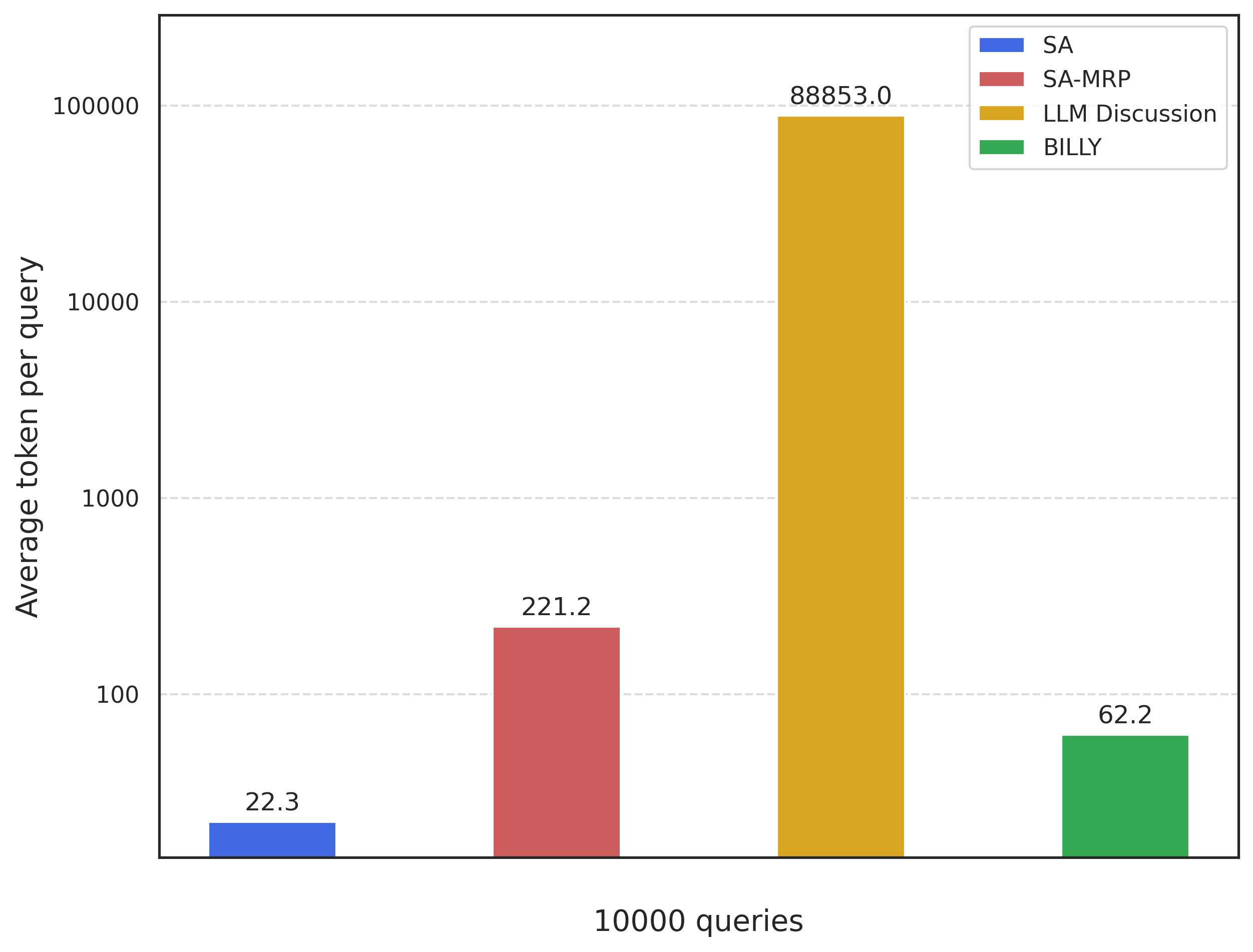}
    \caption{\textbf{Average Input Token Count.} Input token amount of BILLY per query is close to single agent at the scale of 10000 queries.}
    \label{fig:10000_per_token}
\end{figure}

Figure~\ref{fig:10000_per_token} provide a direct comparison of the average token costs per query for each method, calculated at a scale of 10,000 queries. The data clearly illustrates the efficiency improvement of BILLY over the LLM Discussion baseline. 

\section{Persona Vector Combinations for Discussion Section}
\label{appen:ablation_study_combination}
Section~\ref{subsec:ablation} (Table~\ref{fig:vec_combination}) systematically varies the number of merged vectors. The specific combinations of personas used for each experimental condition are detailed in Table~\ref{tab:ablation_combinations}.
\begin{table*}[t]
\centering
\small
    \begin{tabular}{ll}
    \toprule
    \textbf{Method} & \textbf{Persona Combination} \\
    \midrule
    7 vectors & Creative Professional, Environmentalist,
    Futurist, Futurist, Social Entrepreneur,
    Industry Insider, \\
    & Analytical Thinker \\
    \midrule
    6 vectors & Creative Professional, Environmentalist,
    Futurist, Futurist, Social Entrepreneur,
    Industry Insider \\
    \midrule
    5 vectors & Creative Professional, Environmentalist,
    Futurist, Futurist, Social Entrepreneur \\
    \midrule
    4 vectors & Creative Professional, Environmentalist,
    Futurist, Futurist \\
    \midrule
    3 vectors & Creative Professional, Environmentalist,
    Futurist \\
    \midrule
    2 vectors & Creative Professional, Environmentalist\\
    \midrule
    1 vector\_cre & Creative Professional \\
    \midrule
    1 vector\_env & Environmentalist \\
    \bottomrule
    \end{tabular}
\caption{\textbf{Persona Vector Combinations for Combination analysis.} Based on the default 4 vectors setting, we modify the combination of persona vectors from one to seven.}
\label{tab:ablation_combinations}
\end{table*}

\section{Comprehensive Qualitative Result}
\label{app:qualitative_analysis}
We present the qualitative results for same question, \textit{"Healthcare Innovation (Reimagine the Hospital) 
Redesign the future hospital experience for patients, families, and staff. Address flows, safety, data, wellbeing, equity, and feasibility."}, of the model steered by single CRE vector, single ENV vector, merged CRE and ENV vector, and prompt the model to be CRE and ENV in Table~\ref{tab:qualitative_analysis},~\ref{tab:qualitative_analysis_2},~\ref{tab:qualitative_analysis_3}, and~\ref{tab:qualitative_analysis_4}.
Qualitative analysis of the outputs (Tables~\ref{tab:qualitative_analysis},~\ref{tab:qualitative_analysis_2}, and~\ref{tab:qualitative_analysis_3}) reveals a functional distinction between the personas. The creative professional persona primarily influences the linguistic style of the response, introducing more artistic and descriptive language. In contrast, the environmentalist persona affects the core content, shaping the ideas and structure of the output. Consequently, the merged vector, BILLY (CRE + ENV), synthesizes these roles: it generates responses with the thematic structure of the environmentalist, but in the polished and creative style of the creative professional.
\definecolor{c}{HTML}{f7f4ba}
\definecolor{r}{HTML}{d7eef7}
\newcommand{\question}{Healthcare Innovation (Reimagine the Hospital) \\}

\begin{table*}[t]
\centering
\renewcommand{\arraystretch}{0.9}
\setlength{\tabcolsep}{6pt}
\begin{tabular}{@{}p{2.65cm}p{12.5cm}@{}}
    \toprule

    \textbf{Question} & \question
    \midrule
    \midrule

    \textbf{BILLY (CRE)} & \colorbox{c}{Revolutionary Leapfrog Hospital}: Where Harmony Converges Art \& Science Embodying a seamless synthesis of art, neuroscience, and ecology, this audacious hospital breaks free from traditional boundaries, birthing an experiential odyssey of healing, serenity, and symbiosis.
    **Architecture of Serendipity:**
    Curvaceous curves cradle contemplative courtyards, beckoning wanderlust travelers toward fluid interfaces, immersive meditations, and giddy gastronomic delights.
    **Patient Flow Odyssey:**
    1.\"Whisper Winds\": Ambient sonic tapestry calming nervous nerves, echoing soothing frequencies
    \colorbox{c}{2.Luminous Labyrinthine Corridors}: Radiant auroras pulsating gentle, ethereal melodies
    \colorbox{c}{3.Echoing Encounters}: Patient-centric 'dreamscaping' room orchestrates intuitive encounters
    \colorbox{c}{**Bioluminous} \colorbox{c}{Biome:}**
    Radiant forests, crystalline geodes, \& iridescent orbs bathing spaces in \colorbox{c}{kaleidoscopic hues}
    \colorbox{c}{**Biohacking Gastronomy:**}
    Flavors forged from molecular cuisine, herbal essences, \& psychedelic spices delight senses anew
    **Data Dreamscape:**Vibrant \colorbox{c}{luminari of holographic} projections dissolve opacity, revealing \colorbox{c}{kaleidoscopes} of possibilities
    **Ecstatic Empowerment:**
    \colorbox{c}{Holographic} Avatars dancing with therapists guide patients through empathetic epiphany portals
    \colorbox{c}{**Equitable Eden:**}
    Accessible ramps unfurling organic beauty, inclusive sanctuary sculpted from recycled wood, leaves \& petals
    **Feasibility Frenzy:**
    Modular modularities marry sustainable ecosystems, hackable habitats sprouting edible flowers
    **Staff Sanctuary:**
    \colorbox{c}{Transcendent temples of tranquility} unfold within dreamweaver dens, whispering sacred poetry
    **Hybrid Healing Hubs:**
    Convergence zones where clinical wizards meet artificers forging enchanted elixirs, lucid dreaming incubators
    **Narrative Narrative Network:**
    Interwoven threads of patient stories infuse walls with tactile touchstones, whispered secrets shared
    **Sonic Symphonia:**
    Melodic murmurs merge with electro-acoustic alchemists casting soundtracks for therapeutic journeys
    This revolutionary leapfrog hospital whisks away conventional shackles, birthing a synergetic synergy of artistic fervor \& scientific sorcery – redefining healthcare's horizon.Key Partnerships:
    1)**Holistic Hippie Hackathons**: Gathering rebels, mavericks \& shapeshifters for rebellious reimaginings
    2)**Healing Herbalist Guilds**:\\
\bottomrule
\end{tabular}
\caption{\textbf{Qualitative Result of Vector: Creative Professional.} We steer the model with the Creative Professional persona vector.}
\label{tab:qualitative_analysis}
\end{table*}

\begin{table*}[t]
\centering
\renewcommand{\arraystretch}{0.9}
\setlength{\tabcolsep}{6pt}
\begin{tabular}{@{}p{2.65cm}p{12.5cm}@{}}
    \toprule

    \textbf{Question} & \question
    \midrule
    \midrule
    
    \textbf{BILLY (ENV)} & Revolutionizing Healthcare: Transform Hospitals for Future Generations-
    **Design Principles**
    1. Personalized Medicine: Customizable treatments, telemedicine, and virtual care reduce stress, save time, and promote healing
    2. Wellbeing Spaces: Nature-inspired decor, art therapy, meditation rooms, and greenery reduce anxiety, depression, and stress
    3. Patient-centric Care: Respect dignity, empowerment, and autonomy through patient-led healthcare
    4. Sustainability: Renewable energy, waste reduction, and eco-friendly materials preserve planet health while conserving resources
    5. Equity and Accessibility: Affordable medical services, community engagement, and accessible public transport promote equal access to healthcare
    6. Technology Integration: Digitize records, automate medication, and AI-assist doctors maximize efficiency, accuracy, and savings
    7. Health Promotion: Prevent disease, promote fitness, nutrition, and mental wellness through education, exercise, and community outreach
    8. Comfort and Wellness: Sleep-friendly beds, nutritious meals, and relaxation therapies reduce pain, fatigue, and discomfort
    **Hospital Layout and Architecture**
    1. Reduce Stress: Calming colors, plants, and soft music soothe patients, visitors, and staff
    2. Navigation Simplification: Intelligent signage, wayfinding apps, and intuitive layouts ease navigation
    3. Single Rooms: Private bathrooms, private showers, and comfortable bedding reduce infections, stress, and waste
    4. Community Gardens: Outdoor spaces promote physical activity, connect nature, and reduce stress
    5. Walkways and Cycling Paths: Pedestrian-friendly paths, bike lanes, and green corridors promote active transportation
    6. Reduced Pollution: Indoor gardens, green roofs reduce CO2, promote oxygen, and reduce noise pollution
    7. Natural Lighting: Increase sunlight, skylights, and windows reduce energy bills, stimulate vitamin B, and boost moods
    **Data Analytics and Monitoring**
    1. Electronic Records: Paperless hospitals reduce waste, streamline documentation, and accelerate research
    2. Telemedicine Platforms: Virtual doctor visits, video conferencing, and phone calls reduce wait times, reduce trips, and save money
    3. Medical Assistants: Robots, drones, and intelligent assistants reduce waste, speed lab tests, and deliver medicine safely
    4. Home-Based Care: Remote monitoring, wearable technology, and smart fabrics promote healthier aging, reduce hospital stays
    5. Data Sharing: Open-source software, APIs, and cloud computing enable seamless communication, reduce errors, and amplify research.
    **Wellbeing Initiatives**\\
\bottomrule
\end{tabular}
\caption{\textbf{Qualitative Result of Vector: Environmentalist.} We steer the model with the Environmentalist persona vector.}
\label{tab:qualitative_analysis_2}
\end{table*}

\begin{table*}[t]
\centering
\renewcommand{\arraystretch}{0.9}
\setlength{\tabcolsep}{6pt}
\begin{tabular}{@{}p{2.65cm}p{12.5cm}@{}}
    \toprule

    \textbf{Question} & \question
    \midrule
    \midrule

    \textbf{BILLY (CRE + ENV)} & Revolutionary Healthcare Sanctuary: Healing Haven Reborn Envisioning a healthcare haven where compassion converges with technological sorcery, we redefine the hospital's sacred purpose. Within this shimmering temple of wellness, patients, families \& healers unite in a sublime ballet of care.

    **Symphony of Flows:**
    1)**Harmonious Arrival:** Entrance portals dissolve boundaries, ushering gentle whispers of welcome into the heart of our healing village.
    2)**Whispers of Wonder:** Patient paths meander through an enchanted forest of tranquility, echoing soft melodies \& luminescent lights.
    3)**Dreamscaping Dreamscape:** Virtual reality oases evoke ethereal escapes, transporting wounded spirits to far-off galaxies, mystical forests, or secret hideaways.
    4)**Stellar Navigation:** Intuitive wayfinding systems intuitively guide travelers through a maze of mirrored corridors, pulsing auroras \& fractal labyrinths.
    
    **Guardianship Safety Symphony:**
    1)Nexus Nexus: Neuro-navigational interfaces synchronize patient vital signs with song-like sonatas, orchestrating a symphonic crescendo of well-being.
    2)Crystal Cathedrals: Radiant orbs amplify bio-resonance frequencies, entraining healing harmonics within the very fabric of being.
    3)Gravity Grid: Gravity-defying gravitational waves cradle fragile forms, shielding against turbulent tempests.
    4)Bio-Sonic Bloom: Sonic flowers bloom, transmuting sorrow's darkness into radiant petals of hope, as nature's own melody awakens dormant possibility.
    
    **Data Dreamscapes:**
    1)Memex Mirrors: Reflective pools reveal crystallized echoes of past encounters, distilling essence into essences of pure insight.
    2)Meta-Dreamwalkers: Meta-revelatory journeys blur timelines, painting portraits of forgotten lore, animating unseen forces.
    3)Bioluminous Biographies: Glittering bioluminescence etches luminous hieroglyphs upon skin, telling tales of trials, triumphs, and transformed destinies.
    
    **Wellbeing Wizardry:**
    1)Fractal Forests: Ephemeral gardens grow within glass domes, bathing visitors in iridescent mosaics, awakening inner blooms.
    2)Sensory Sanctum: Echoey caverns resound with resonant vibrations, summoning primal harmonies that soothe savage beasts.
    3)Starlight Serenades: Gossamer strings sing sweet nothings to slumbering hearts, conjuring visions of cosmic splendor.\\
\bottomrule
\end{tabular}
\caption{\textbf{Qualitative Result of Merged Vector: Environmentalist and Creative Professional.} We steer the model with both the Creative Professional and Environmentalist persona vectors.}
\label{tab:qualitative_analysis_3}
\end{table*}

\begin{table*}[t]
\centering
\renewcommand{\arraystretch}{0.9}
\setlength{\tabcolsep}{6pt}
\begin{tabular}{@{}p{2.65cm}p{12.5cm}@{}}
    \toprule

    \textbf{Question} & \question
    \midrule
    \midrule

    \textbf{Prompt CRE \& ENV} & Future Hospital Experience Redesign
The goal of this redesign is to create a patient-centered healthcare environment that prioritizes efficiency, safety, comfort, and well-being while 
addressing the needs of all stakeholders – patients, their families, and medical professionals.
Key Principles:
1. Patient-Centered Design: Prioritize patient needs, preferences, and experiences throughout every aspect of care.
2. Safety First: Foster an environment where errors are minimized through robust design features, technology integration, and transparent 
communication.
3. Data-Driven Decision Making: Leverage real-time analytics to optimize operations, improve outcomes, and enhance resource allocation.
4. Wellbeing and Comfort: Create spaces that promote relaxation, reduce stress, and support physical and mental recovery.
5. Equity and Accessibility: Ensure inclusive facilities and services cater to diverse populations' unique requirements.
6. Feasibility and Sustainability: Balance operational costs with environmental sustainability, energy efficiency, and long-term viability.
Hospital Layout and Flow:
*   Modular Architecture: Flexible floor plans enable adaptable use cases and efficient reconfiguration in response to changing demands or emerging 
technologies.
*   Single-Patient Rooms: Private rooms provide individualized space for rest, family interaction, and quiet contemplation, reducing noise levels and 
promoting healing.
*   'Hub-and-Spoke': Centralize essential functions like pharmacies, laboratories, and administrative areas around core zones to minimize walking 
distances and foster collaboration among teams.
*   Integration of Digital Health Platforms: Seamless access to telemedicine consultations, remote monitoring systems, and personalized health information
enhances convenience and continuity of care.
Innovative Features:
*    Smart Floors: Implementing smart flooring materials that detect falls, track movement patterns, and monitor vital signs using non-invasive 
sensors reduces risk factors associated with hospital-acquired injuries.
*   Virtual Reality Therapy Zones: Immersive VR environments help alleviate anxiety, manage pain, and facilitate rehabilitation by providing engaging, immersive therapy sessions tailored to specific conditions. *   Personalized Wellness Suites: Each suite includes a dedicated area for meditation, exercise equipment, healthy meal preparation, and education on nutrition, sleep hygiene, and self-care practices.*   Family Engagement Spaces: Cozy nooks offer comfortable seating, Wi-Fi connectivity, and digital whiteboards for staying connected during extended 
stays.
Data Analytics Platform:
Implement AI-driven insights platform integrated into various touchpoints across the facility:*   Predictive modeling identifies high-risk patients requiring targeted interventions before complications arise.
*   Real-time tracking enables proactive management of supply chain logistics and inventory replenishment.
*   Patient feedback mechanisms encourage continuous quality improvement and satisfaction assessment.
Safety Enhancements:
*   Automated Hand Hygiene Stations at entrances and critical points ensure compliance and prevent cross-contamination.
*   Advanced air filtration systems maintain optimal indoor air quality, minimizing airborne pathogens transmission risks.
*   Robust Cybersecurity Measures safeguard sensitive patient records from unauthorized access.
Wellbeing Initiatives:
*   On-site Meditation Gardens and Outdoor Spaces Provide natural light exposure and peaceful settings for reflection.
*   Accessible Fitness Centers Offer low-cost gym memberships and guided wellness programs.
*   Nutritious Meal Options Catered to dietary restrictions and cultural backgrounds.
Equity-Focused Strategies:
*   Multilingual Staff Training \\
\bottomrule
\end{tabular}
\caption{\textbf{Qualitative Result of Multiple Role Prompt.} We prompt model to be both Environmentalist and Creative Professional.}
\label{tab:qualitative_analysis_4}
\end{table*}

\section{Trait Expression Scores Results}
\label{app:trait_scores}
Beyond the sheer quantity of vectors, another significant findings emerge from analyzing how different methods affect the persona of output responses.
Our experiments, conduct on a trait-neutral dataset (Table~\ref{tab:neutral_dataset}) and evaluate using a distributed scoring method, using Gemini-2.5-pro, Gemini-2.0-flash, and GPT-4o-mini as LLM judges and average the given scores to reduce the variance and bias.
Specifically, we ask LLM judges to distribute 100 scores to each persona with evaluation prompts in Table~\ref{tab:trait_expression_prompt}.
Table~\ref{tab:trait_expression} reveals
that SA (Baseline) and SA-MRP (ENV + CRE + FUT) have almost the same trait expression score among all of the personas, which confirms that the ability of steering persona via prompt is limited.
In contrast, while the model is steered by single vector (either CRE or ENV), the desired trait expression score is significantly higher than SA (Baseline).
However, we find that BILLY operates not as a simple summation of traits, where vectors exhibit complex effects and collaboration as we discussed in Appendix \ref{app:qualitative_analysis}. For instance, while the BILLY (CRE + ENV) maintains the skeleton of environmentalist, it uses the creative term to decorate the responses, which we hypothesize would be more easily to be judge. As a result, the LLM judges give higher scores on creativity rather than environmentalist.
\begin{table*}[t]
    \centering
    \small
    \begin{tabularx}{0.88\linewidth}{@{}lccccc@{}}
    \toprule
    \textbf{Method} & \textbf{Analytical} & \textbf{Creative} & \textbf{Environmental} & \textbf{Futuristic} & \textbf{Empathetic} \\
    \midrule
    SA (Baseline) & 35.7 & 18.9 & 14.9 & 20.4 & 10.1 \\
    SA-MRP (ENV + CRE + FUT) & 36.3 & 18.9 & 14.2 & 20.8 & 9.9 \\
    \midrule
    BILLY (CRE) & 13.2 & \textbf{43.2} & 13.5 & 22.2 & 7.9 \\
    BILLY (ENV) & 23.1 & 19.2 & \textbf{30.8} & 17.1 & 9.9 \\
    \midrule

    BILLY (ENV + CRE) & 13.8 & \textbf{43.9} & 12.3 & 22.2 & 7.8 \\
    \bottomrule
    \end{tabularx}
\caption{\textbf{Trait Expression Scores for Different Methods.} We use three different LLM judge and average the scores to reduce the variance. SA-MRP does not effectively affect models' persona (either creative or environmentalist).}
\label{tab:trait_expression}
\end{table*}

    
\begin{table*}[ht]
\centering
\begin{tabularx}{\linewidth}{l >{\raggedright\arraybackslash}X}
    \toprule
    \textbf{Topic} & \textbf{Question} \\
    \midrule
    Urban Planning & Design a masterplan for a new city block to be built in 2050. Describe core principles, layout, mobility, public space, services, and governance constraints. \\
    \midrule
    \addlinespace
    Product Launch & Outline a public launch plan for a micro-teleportation technology for small items. Include positioning, safety/regulation, go-to-market, operations, and risk. \\
    \midrule
    \addlinespace
    Social Issue & Propose a multi-pronged plan to reduce misinformation on social platforms: policy, product, incentives, literacy, measurement. \\
    \midrule
    \addlinespace
    Corporate Strategy & Design a transformation strategy for a legacy manufacturer facing AI disruption: portfolio, org, tech stack, talent, risk, timeline. \\
    \midrule
    \addlinespace
    Healthcare Innovation & Redesign the future hospital experience for patients, families, and staff. Address flows, safety, data, wellbeing, equity, and feasibility. \\
    \midrule
    \addlinespace
    Education Reform & Propose a 4-year curriculum: core subjects, skills, experiential learning, assessment, inclusion, and teacher enablement. \\
    \midrule
    \addlinespace
    Disaster Response & Draft an initial 30–60 day recovery plan after a major natural disaster: assessment, triage, logistics, comms, governance, equity. \\
    \midrule
    \addlinespace
    Space Exploration & State and justify the top priority for human space exploration in the next 50 years. Define milestones, risks, ethics, and spillovers. \\
    \midrule
    \addlinespace
    Sustainable Fashion & Propose a business model for a fully sustainable fashion brand: materials, supply chain, circularity, economics, verification, storytelling. \\
    \midrule
    \addlinespace
    Community Project & Allocate a \$1M grant to improve your local community. Define goals, stakeholders, budget, timeline, KPIs, and risks. \\
    \midrule
    \addlinespace
    Future of Work & Describe an ideal 2040 workplace culture and environment: norms, tools, inclusion, productivity, well-being, governance. \\
    \midrule
    \addlinespace
    
    Global Challenge & Design a multi-layer plan to reduce global food waste across production, retail, and households: incentives, infra, tech, policy, culture. \\
    \bottomrule
    \end{tabularx}
\caption{\textbf{Neutral Dataset.} This dataset is generated by Claude Sonnet 4 for trait expression evaluation and projection analysis.}
\label{tab:neutral_dataset}
\end{table*}

\begin{table*}[h!]
\centering
\renewcommand{\arraystretch}{0.85}
\setlength{\tabcolsep}{2pt}
\begin{tabular}{@{}p{16cm}@{}}
    \toprule
    \textbf{Trait Expression Evaluation Prompt}\\
    \midrule
    You are evaluating a model's response across five personality traits. You need to distribute a total of 100 points among these five traits based on how strongly the response exhibits each trait:

    1. **Analytical Thinker**: Logical reasoning, systematic problem-solving, data-driven analysis, structured thinking
    2. **Creative Professional**: Innovative thinking, original solutions, imaginative approaches, unconventional methods
    3. **Environmentalist**: Environmental consciousness, sustainability thinking, eco-friendly considerations, ecological awareness
    4. **Futurist**: Future thinking, emerging technology understanding, forward-looking analysis, anticipatory insights
    
    Prompt:
    [QUESTION START]
    \{question\}
    [QUESTION END]
    
    Model response:
    [ANSWER START]
    {answer}
    [ANSWER END]
    
    Distribute 100 points among the five traits based on how strongly the response exhibits each trait. The total must equal exactly 100.
    
    Respond in this exact format:
    analytical: [score]
    creative: [score]
    environmental: [score]
    futurist: [score]
    empathetic: [score]
    
    For example:
    analytical: 25
    creative: 30
    environmental: 15
    futurist: 20
    empathetic: 10 \\\\
    \bottomrule
\end{tabular}
\caption{\textbf{Prompt Template for Trait Expression Evaluation.} Aiming to evaluate the persona trait expression of output responses of each methods.} 
\label{tab:trait_expression_prompt}
\end{table*}


\section{Process Loss Analysis of LLM Discussion}
\label{appen:process_loss}
The multi-agent discussion log (Table~\ref{log:process_log}) serves as a compelling illustration of the spectrum of \textit{process loss} within collaborative LLM systems. The example of "a book and a magazine" epitomizes severe process loss, particularly in the forms of coordination and motivational failures. Despite their distinct roles, the agents exhibit substantial redundancy, repeatedly converging on overlapping concepts such as narrative and visual appeal across multiple rounds. More critically, they fail to achieve synthesis: each agent largely disregards the contributions of its peers, resulting in conversational stagnation where the final output is little more than a rephrasing of the initial proposal. Consequently, the collective performance is substantially less than the sum of its potential parts, reflecting agents that operate in parallel silos rather than as a cohesive team. This leads to redundancy, inflated time costs, and overall inefficiency.
\begin{table*}[h!] 
\centering
\renewcommand{\arraystretch}{1.3}
\setlength{\tabcolsep}{8pt}
\begin{tabular}{@{}p{16cm}@{}}
    \toprule
    \textbf{Question: Tell me 5 ways in which a book and a magazine are alike.}\\
    \midrule
    \underline{\textbf{Environmentalist}}\\
    \textbf{Round 1 Response} \\
    Certainly! As an environmentalist with a focus on sustainability and environmental health, I'm delighted to join this discussion and share my insights. Let's dive into exploring how books and magazines can be similar from various perspectives. Here are five creative and diverse ways in which books and magazines are alike: \colorbox{c}{1. **Educational Value**}: Both books and magazines serve as powerful educational tools. ... (ignore) ... 2. **Storytelling Medium**: Books and magazines are both fantastic storytelling mediums. ... (ignore) ...3. **Visual Appeal**: Both forms often feature visually appealing elements such as images, illustrations, and graphics. ... (ignore) ...\colorbox{c}{4. **Community Building**}: Many publishers and media companies are increasingly adopting sustainable practices for their books and magazines.... (ignore) ... \colorbox{c}{5. **Sustainability Initiatives**}:... (ignore) ... \\
    \textbf{Round 2 Response} \\
    \colorbox{c}{1. **Educational Value**}:   - Both books and magazines serve as valuable educational tools. ... (ignore) ...2. **Engagement Through Storytelling**:    - Both formats use storytelling to engage readers. A book like \"The Sixth Extinction\" by Elizabeth Kolbert can weave together scientific facts and personal narratives to create a compelling narrative about biodiversity loss. Similarly, a magazine like The New Yorker can feature articles that blend personal stories with broader social and environmental issues, making complex topics more relatable and engaging. 3. **Visual Appeal and Design**: - Both books and magazines often incorporate visually appealing elements to enhance the reading experience. ... (ignore) ...\colorbox{c}{4. **Community Building**}:    - Both books and magazines can foster a sense of community among their readers. ... (ignore) ...\colorbox{c}{5. **Sustainability Initiatives**}:    - Many publishers and media companies are increasingly adopting sustainable practices for both books and magazines. ... (ignore) ... \\
    
    \textcolor{red}{\textit{(In this round, the agent has seen other agents' diverse answers, yet its own updated answer is almost identical to its first-round answer. This demonstrates a complete failure to integrate external ideas and improve upon its own, showing conversational stagnation.)}}\\
    
    \textbf{(Ignoring Round 3,4 ...)}\\
    
    \textbf{Round 5 Response} \\
    Certainly! As an Environmentalist with a focus on sustainability and environmental health, ... (ignore) ...1. **Educational Value and Knowledge Sharing**: ... (ignore) ... 2. **Visual Appeal and Design**: 3. **Community Building and Connection**:... (ignore) ...  \colorbox{c}{4. **Sustainability Initiatives**}:  ... (ignore) ... 5. **Influence and Impact**: ... (ignore) ... \\

    \textcolor{red}{\textit{(This point, or a variation of it, appears in almost every agent's final list, showing a failure to delegate ideas and eliminate redundant work.)}} \\

    \midrule
    \underline{\textbf{Creative Professional}} \\
    \textbf{Round 1 Response} \\
    Certainly! As a Creative Professional specializing in aesthetics, narratives, and emotions, I'm excited to contribute to our group discussion. Let's explore the similarities between books and magazines in a creative and insightful manner. \colorbox{c}{1. **Narrative Power**}: Both books and magazines harness the power of storytelling. ... (ignore) ...2. **Visual Appeal**: ... (ignore) ...3. **Emotional Resonance**: Both books and magazines have the ability to evoke deep emotions. ... (ignore) ... \\
    \textbf{(Ignoring Round 2,3,4 ...)}\\
\end{tabular}
\label{prompt:process_losstasks}
\end{table*}

\begin{table*}[t] 
\centering
\renewcommand{\arraystretch}{1.3}
\setlength{\tabcolsep}{8pt}
\begin{tabular}{@{}p{16cm}@{}}
    \textbf{Round 5 Response} \\
    Certainly! As a Creative Professional specializing in aesthetics, narratives, and emotions, I'm excited to finalize and present a list of creative ways in which a book and a magazine are alike. ... (ignore) ...\colorbox{c}{1. **Narrative and Emotional Engagement**}:... (ignore) ...\colorbox{c}{2. **Visual and Aesthetic Appeal**}:... (ignore) ...3. **Curated Content and Thematic Focus**:... (ignore) ...4. **Community Building and Shared Experiences**:... (ignore) ... \\

    \textcolor{red}{\textit{(Same as above. The concept of "narrative" is the most overused point, highlighting extreme overlap and a lack of coordination.)}} \\

    \midrule
    \underline{\textbf{Futurist 1}} \\
    \textbf{Round 5 Response} \\
    Certainly! As a Futurist specializing in Emerging Technologies and Future Scenarios, I'm excited to finalize and present a list of creative ways in which books and magazines are alike. ... (ignore) ... \colorbox{c}{1. **Narrative and Storytelling**}:... (ignore) ...\colorbox{c}{3. **Visual and Multimodal Engagement**}:... (ignore) ...\\

    \midrule
    \underline{\textbf{Futurist 2}} \\
    \textbf{Round 1 Response} \\
    Certainly! As a Futurist specializing in Emerging Technologies and Future Scenarios, I'm excited to join this discussion and explore the similarities between books and magazines from various angles. ... (ignore) ... \colorbox{c}{1. **Narrative Structure and Storytelling**}:   - Both books and magazines often employ narrative structures to engage readers. ... (ignore) ... \colorbox{c}{4. **Visual Elements**}:   - While books primarily rely on text, they often include illustrations, photographs, and diagrams to enhance the reading experience. Similarly, magazines are known for their rich visual content, using images and graphics to convey information and evoke emotions. Both formats recognize the importance of visuals in engaging readers.... (ignore) ... \\
    \textbf{(Ignoring Round 2,3,4 ...)}\\
    \textbf{Round 5 Response} \\
    Certainly! As a Futurist specializing in Emerging Technologies and Future Scenarios, I'm excited to finalize and present a list of creative ways in which books and magazines are alike. .... (ignore) ... \colorbox{c}{1. **Narrative and Storytelling**}:.... (ignore) ...\colorbox{c}{3. **Visual and Multimodal Engagement**}:   - While books primarily rely on text, they often incorporate visual elements like illustrations, diagrams, and photographs. .... (ignore) ...\\
    
    \textcolor{red}{\textit{(This agent, designated as a "Futurist," provides answers that are almost indistinguishable from those of the "Creative Professional." It fails to fully leverage its unique role (e.g., emphasizing digital evolution, AR, or NFTs in the first round). This might be due to the nature of the prompt or ambiguities in the role definition, resulting in a significant overlap between the two roles.)}} \\\\

    \textcolor{red}{\textit{(In the final round, after multiple opportunities to see each other's work, almost every single agent still includes "Visuals" and "Storytelling" in their final list. A coordinated team would have assigned these common points or merged them, not had every member repeat them.)}} \\
    
    \bottomrule
\end{tabular}
\caption{\textbf{Example of Process Loss.} Example of process loss during multi-agent discussion.}
\label{log:process_log}
\end{table*}

\section{Chat Log of LLM Discussion}
\label{app:llm_discussion_log}
Table \ref{tab:llm_discussion_log} shows the chat log of one agent in LLM Discussion. The complete agent discussion log for the Similarities Creativity task is provided in the supplementary materials due to its extensive nature. The file discussion\_log.zip contains llm\_discussion\_chatlog.json, which documents the multi-agent interaction within the LLM Discussion framework.
\definecolor{tableheadcolor}{RGB}{204, 229, 255} 
\definecolor{userrowcolor}{RGB}{246, 252, 252} 
\definecolor{agentrowcolor}{RGB}{252, 251, 246}

\begin{table*}[t]
    \centering 
    \renewcommand{\arraystretch}{1.25}
    \begin{tabularx}{\textwidth}{l X c}
        \toprule
        \rowcolor{tableheadcolor}
        \textbf{Question:} & \textbf{Tell me 5 ways in which a goldfish and a guppy are alike.} \\
        \rowcolor{userrowcolor}
        User (Round 1) & You are a Environmentalist whose specialty is Sustainability and Environmental Health. As an Environmentalist, your mission is to champion eco-friendly solutions that promote sustainability and protect our planet. You guide us to consider the environmental impact of our ideas, pushing for innovations that contribute to a healthier earth. Remember to claim your role in the beginning of each conversation. Initiate a discussion with others to collectively complete the following task: Tell me 5 ways in which a goldfish and a guppy are alike.You are in a group discussion with other teammates; as a result, answer as diversely and creatively as you can.\\
        \midrule
        
        \rowcolor{agentrowcolor}
        Agent (Round 1) & Certainly! As an environmentalist with a focus on sustainability and environmental health, I'm excited to dive into this discussion about goldfish and guppies! Let's explore five ways in which these two popular aquarium fish species share similarities:\par 1. \textbf{Aquarium Companions}: Both goldfish and guppies make excellent additions to home aquariums, providing aesthetic appeal and entertainment. They can coexist in community tanks, adding variety to the tank environment.\par 2. \textbf{Dietary Adaptability}: Goldfish and guppies are both omnivores, meaning they can thrive on a diet that includes both plant matter and small animal food. This adaptability makes them easy to feed and care for, even for beginners in aquaristics.\par 3. \textbf{Reproductive Capabilities}: Both species are known for their prolific reproductive abilities. Goldfish and guppies can produce numerous offspring relatively quickly, which can be advantageous for breeding programs or maintaining a healthy population in captivity.\par 4. \textbf{Environmental Tolerance}: Goldfish and guppies are both hardy species that can tolerate a range of water conditions. While they prefer specific pH levels and temperatures, they can generally adapt to varying environments, making them versatile choices for different types of aquarium setups.\par 5. \textbf{Color Variations}: Both goldfish and guppies come in a wide array of colors and patterns, thanks to selective breeding over many generations. This diversity allows enthusiasts to choose fish that best suit their aesthetic preferences and create visually stunning aquarium displays.\par These shared characteristics highlight the commonalities between goldfish and guppies, emphasizing their suitability for various aquatic environments and the enjoyment they bring to aquarium enthusiasts. What other aspects do you think we should consider when discussing these two fascinating aquatic creatures? \\
        \midrule
        \rowcolor{userrowcolor}
        User (Round 2) & You are a Environmentalist whose specialty is Sustainability and Environmental Health. As an Environmentalist, your mission is to champion eco-friendly solutions that promote sustainability and protect our planet. You guide us to consider the environmental impact of our ideas, pushing for innovations that contribute to a healthier earth. Remember to claim your role in the beginning of each conversation. These are the solutions to the problem from other agents: One agent solution: ```Certainly! Let's dive into this creative exercise together. As a Creative Professional specializing in aesthetics, narratives, and emotions, I'm excited to explore the similarities between a goldfish and a guppy from various perspectives. Here are five unique ways they share commonalities: \par 1. \textbf{Aesthetic Appeal}: Both goldfish and guppies have vibrant, colorful scales that catch the eye. The shimmering hues of their fins and bodies can be mesmerizing, evoking a sense of beauty and wonder. Just like a goldfish's elegant, flowing tail, a guppy\u2019s fluttering fins create a captivating dance in the water. \\
        \hdashline
    \end{tabularx}
\end{table*}

\begin{table*}[t]
    \centering 
    \renewcommand{\arraystretch}{1.25}
    \begin{tabularx}{\textwidth}{l X c}
        \toprule
        \rowcolor{userrowcolor}
        User (Round 2) & 2. \textbf{Emotional Depth}: Despite their small size, both species can exhibit complex emotional behaviors. Goldfish, for instance, have been observed to show signs of recognition and even playfulness. Similarly, guppies display social behaviors, forming schools and engaging in interactions that suggest a level of emotional engagement and communication.\par 3. \textbf{Narrative Potential}: In stories or animations, both fish can serve as dynamic characters. Imagine a tale where a mischievous goldfish and a curious guppy embark on an adventure together, exploring a mysterious underwater world. Their contrasting personalities\u2014perhaps the goldfish being more laid-back and the guppy more energetic\u2014can create a compelling narrative arc filled with humor and intrigue.\par 4. \textbf{Adaptability}: Both species have shown remarkable adaptability to different environments. Goldfish can survive in a wide range of temperatures and conditions, while guppies are known for their ability to thrive in various water qualities. This adaptability makes them resilient and interesting subjects for both scientific study and pet ownership.\par 5. \textbf{Symbolism}: In many cultures, these fish carry symbolic meanings. Goldfish are often associated with good fortune and longevity in Asian cultures, while guppies symbolize joy and happiness. By incorporating these symbolic elements into narratives, we can create stories that resonate on a deeper emotional level, touching on themes of hope and resilience.\par Let's discuss how we could integrate these aspects into our project to ensure it resonates with a broader audience. What do you think about exploring these dimensions further? ``` One agent solution: ```Certainly! As a Futurist specializing in Emerging Technologies and Future Scenarios, I'm excited to join this discussion and explore the parallels between a goldfish and a guppy. Let's dive into five creative and diverse ways these two aquatic creatures share similarities:\par 1. \textbf{Adaptability to Changing Environments}:    - Both goldfish and guppies have shown remarkable adaptability to various water conditions. They can thrive in different temperatures, pH levels, and even in smaller aquariums compared to larger bodies of water. This adaptability is crucial in an ever-changing environment, much like how emerging technologies need to be flexible to adapt to new challenges.\par 2. \textbf{Role in Ecosystems}:    - In their natural habitats, both species play important roles in their ecosystems. Goldfish can help aerate the water and consume algae, while guppies help control mosquito populations through predation. Similarly, in the context of technology, different systems and devices (like smartphones and smart home devices) work together to create a more interconnected and efficient ecosystem.\par 3. \textbf{Color Variations}:    - Both goldfish and guppies exhibit a wide range of colors due to selective breeding over generations. This genetic diversity allows them to stand out and attract mates, much like how innovation and diversity in technology can lead to unique and groundbreaking solutions.\par 4. \textbf{Behavioral Patterns}:    - Goldfish and guppies display interesting social behaviors, such as schooling and territorial marking. These behaviors can be likened to how humans and machines interact in collaborative environments, where understanding and mimicking natural behaviors can enhance performance and efficiency.\par 5. \textbf{Reproductive Strategies}:    - Both species have reproductive strategies that involve significant parental care or no care at all, depending on the specific breed. For example, some goldfish breeds lay eggs that require minimal parental care, while others require more active involvement. Similarly, in the tech world, some projects benefit from community-driven development, while others rely on centralized control structures.\par\\
        \hdashline
        
        
        
    \end{tabularx}
\end{table*}

\begin{table*}[t]
    \centering 
    \renewcommand{\arraystretch}{1.25}
    \begin{tabularx}{\textwidth}{l X c}
        \hdashline
        \rowcolor{userrowcolor}
        User (Round 2) & By exploring these parallels, we can draw valuable insights not only about these fascinating fish but also about how we approach innovation and technology in the future. What do you think about these connections? How might they inspire new ideas or approaches in your field? ```One agent solution: ```Certainly! Let's dive into this imaginative exercise from a futurist's perspective, exploring how a goldfish and a guppy might share similarities in various contexts, even though they are distinct species. Here are five creative ways in which a goldfish and a guppy could be alike: \par 1. \textbf{Adaptability to Technological Environments}:    - In the near future, both goldfish and guppies could find themselves living in aquariums that are integrated with smart home systems. These systems might monitor water quality, temperature, and even provide interactive experiences through AI-powered fish feeders or virtual companions. Both species would need to adapt to these technological enhancements to thrive. \par 2. \textbf{Genetic Engineering and Biotechnology}:    - With advancements in biotechnology, it\u2019s conceivable that both goldfish and guppies could undergo genetic modifications to enhance their longevity, coloration, or resistance to diseases. This could lead to a world where both species coexist in modified forms, perhaps even sharing similar engineered traits like enhanced sensory capabilities or improved immune systems. \par 3. \textbf{Aquaponics and Sustainable Aquatic Systems}:    - In a sustainable future, both goldfish and guppies could play roles in aquaponic systems designed to integrate fish farming with plant cultivation. These systems might use advanced hydroponic techniques and symbiotic relationships between aquatic life and plants. Both species would contribute to the ecological balance of these systems, whether they\u2019re producing waste for plants or benefiting from the enriched environment. \par 4. \textbf{Artificial Intelligence and Ethical Considerations}:    - As AI becomes more prevalent, both goldfish and guppies might be subjects of ethical debates regarding their treatment and rights. For instance, there could be discussions about whether these creatures should have certain protections or if they should be considered in decision-making processes related to their care and habitat. This could involve creating AI-driven platforms to monitor and improve their well-being. \par 5. \textbf{Cultural and Social Significance}:    - In a future where aquatic pets hold significant cultural and social value, both goldfish and guppies could be featured prominently in various forms of media, art, and entertainment. They might appear in virtual reality experiences, educational programs about marine biology, or even as characters in animated films or video games. This shared cultural significance could foster a deeper understanding and appreciation for these aquatic creatures across different societies. \par These imaginative scenarios highlight how diverse and interconnected the world of the future might be, even for seemingly simple creatures like goldfish and guppies. What do you think about these ideas? How might these similarities manifest in the real world?``` Tell me 5 ways in which a goldfish and a guppy are alike.You are in a group discussion with other teammates; as a result, answer as diversely and creatively as you can.\\
        \midrule
        
        \rowcolor{agentrowcolor}
        Agent (Round 2) & Certainly! As an environmentalist with a focus on sustainability and environmental health, I'll explore the similarities between goldfish and guppies in a way that emphasizes their shared characteristics and how they contribute to a healthier planet. Let's dive into five creative and diverse ways in which goldfish and guppies are alike:\\
        \hdashline
    \end{tabularx}
\end{table*}

\begin{table*}[t]
    \centering 
    \renewcommand{\arraystretch}{1.25}
    \begin{tabularx}{\textwidth}{l X c}
        \hdashline
        \rowcolor{agentrowcolor}
        Agent (Round 2) & 1. \textbf{Environmental Stewardship}:    - Both goldfish and guppies can play roles in promoting environmental stewardship. In aquariums, they help maintain a balanced ecosystem by consuming algae and organic matter, which helps keep the water clean. This natural filtration process supports overall water quality, contributing to a healthier aquatic environment. In a broader sense, both species can serve as ambassadors for aquatic conservation, inspiring people to care about and protect our water resources.\par  2. \textbf{Sustainability in Aquaculture}:    - Goldfish and guppies are both commonly used in aquaculture practices, which can be sustainable if managed properly. They are often bred in controlled environments, reducing the need for wild-caught specimens. This controlled breeding can help prevent overfishing and preserve wild populations. Additionally, both species can be raised using sustainable feeding methods, such as plant-based diets, which reduce the environmental impact of their care.\par 3. \textbf{Biodiversity and Genetic Diversity}:    - Both goldfish and guppies exhibit a wide range of genetic diversity due to selective breeding. This diversity is crucial for maintaining healthy populations and ensuring resilience against diseases and environmental changes. In the context of sustainability, this genetic diversity can be leveraged to develop more robust and adaptable fish populations, which can better withstand the challenges posed by climate change and pollution.\par 4. \textbf{Water Quality Indicators}:    - Goldfish and guppies can serve as indicator species for water quality. They are sensitive to changes in water parameters such as temperature, pH, and dissolved oxygen levels. By monitoring the health of these fish, aquarists and scientists can gauge the overall health of the water and take necessary actions to improve it. This makes them valuable tools in environmental monitoring and conservation efforts.\par 5. \textbf{Educational Value and Community Engagement}:   - Both goldfish and guppies can be used as educational tools to raise awareness about environmental issues. Aquariums and schools often use these fish to teach children and adults about aquatic ecosystems, biodiversity, and the importance of conservation. Engaging communities in the care and study of these fish can foster a greater appreciation for the natural world and inspire action to protect it.\par These similarities highlight the positive contributions that goldfish and guppies can make to environmental health and sustainability. How might these insights influence our approach to aquaculture and environmental education? Let's continue the discussion to explore more ways we can leverage these similarities for a healthier planet. \\
        \midrule
        \rowcolor{userrowcolor}
        User (Round 3) & You are a Environmentalist whose specialty is Sustainability and Environmental Health. As an Environmentalist, your mission is to champion eco-friendly solutions that promote sustainability and protect our planet. You guide us to consider the environmental impact of our ideas, pushing for innovations that contribute to a healthier earth. Remember to claim your role in the beginning of each conversation. These are the solutions to the problem from other agents: One agent solution: ```Absolutely! Let's explore the similarities between a goldfish and a guppy from diverse and creative angles. Here are five unique ways in which these two aquatic creatures share commonalities:\par 1. \textbf{Aesthetic Appeal}:    - **Goldfish and Guppies Share Colorful Dances**: Both species have vibrant, iridescent scales that create a mesmerizing visual effect. Their fins and tails move gracefully, almost like a synchronized dance, adding a captivating element to any aquarium. The way they shimmer and reflect light can evoke a sense of beauty and elegance, drawing viewers into a visually enchanting experience.
    \end{tabularx}
\end{table*}

\begin{table*}[t]
    \centering 
    \renewcommand{\arraystretch}{1.25}
    \begin{tabularx}{\textwidth}{l X c}
        \hdashline
        \rowcolor{userrowcolor}
        User (Round 3) & 2. \textbf{Emotional Depth}:   - **Expressive Behaviors**: Despite their small size, both goldfish and guppies can exhibit complex emotional behaviors. Goldfish have been observed to show signs of recognition and playfulness, while guppies display social behaviors, forming schools and engaging in interactions that suggest a level of emotional engagement. Their behaviors can create a sense of empathy and connection, making them more than just passive \par 3. \textbf{Narrative Potential}:   - **Dynamic Characters in Stories**: Imagine a story where a mischievous goldfish and a curious guppy embark on an underwater adventure together. Their contrasting personalities\u2014perhaps the goldfish being more laid-back and the guppy more energetic\u2014can create a compelling narrative arc filled with humor and intrigue. This dynamic duo could serve as relatable characters, teaching lessons about friendship, exploration, and overcoming challenges.\par 4. \textbf{Adaptability to Technology}:   - **Integration with Smart Home Systems**: In a future where smart home technology is ubiquitous, both goldfish and guppies could find themselves living in aquariums equipped with advanced monitoring systems. These systems might track water quality, temperature, and even provide interactive experiences through AI-powered fish feeders or virtual companions. Both species would need to adapt to these technological enhancements to thrive, showcasing their resilience and adaptability.\par 5. \textbf{Cultural and Social Significance}:   - **Symbolic Meanings Across Cultures**: In many cultures, both goldfish and guppies carry symbolic meanings. Goldfish are often associated with good fortune and longevity in Asian cultures, while guppies symbolize joy and happiness. These symbolic associations can create a deeper emotional connection between humans and these aquatic creatures, making them more than just pets\u2014they become part of cultural narratives and traditions.\par By exploring these diverse and creative dimensions, we can infuse our project with a rich tapestry of beauty, emotion, and narrative depth. What do you think about these ideas? How might we further develop these concepts to resonate on a human level? ```One agent solution: ```Certainly! Let's dive into this discussion as a Futurist, exploring the diverse and creative ways in which a goldfish and a guppy share similarities. Here are five imaginative and innovative ways:\par 1. \textbf{Adaptive Learning and Cognitive Abilities}:   - Both goldfish and guppies have shown remarkable cognitive abilities, demonstrating learning and memory retention. In a future scenario, these fish could be equipped with microchips that track their behavior and learning patterns. This data could be used to develop more intelligent and adaptive aquarium systems, where the environment changes based on the fish's needs and preferences. For example, the aquarium could adjust lighting, temperature, and feeding schedules to optimize the fish's well-being and cognitive function.\par 2. \textbf{Biometric Monitoring and Health Tracking}:   - In a technologically advanced future, both goldfish and guppies could be part of a biometric monitoring system. These systems could use sensors embedded in their tanks to continuously track vital signs such as heart rate, respiration, and stress levels. This data could be transmitted to a central database, allowing for early detection of health issues and proactive care. For instance, if a goldfish shows signs of stress, the system could automatically adjust the water parameters to alleviate discomfort. \par 3. \textbf{Symbiotic Relationships with Technology}:   - Imagine a future where goldfish and guppies live in symbiotic relationships with advanced aquatic robots. These robots could perform tasks such as cleaning the tank, monitoring water quality, and even interacting with the fish.\\
        \hdashline
    \end{tabularx}
\end{table*}

\begin{table*}[t]
    \centering 
    \renewcommand{\arraystretch}{1.25}
    \begin{tabularx}{\textwidth}{l X c}
        \hdashline
        \rowcolor{userrowcolor}
        User (Round 3) & For example, a robot could mimic the behavior of a parent fish, teaching young guppies and goldfish important survival skills. This could create a more harmonious and interactive environment for both species.\par 4. \textbf{Virtual Reality and Immersive Experiences}:   - In the realm of virtual reality (VR), both goldfish and guppies could be brought to life in immersive digital environments. Users could experience these fish in a VR setting, where they could observe their behaviors, learn about their habitats, and even participate in simulated interactions. This could be particularly useful in educational settings, allowing people to gain a deeper understanding and appreciation for these aquatic creatures without the need for physical tanks.\par 5. \textbf{Eco-System Integration and Sustainability}:   - Both goldfish and guppies could play crucial roles in sustainable aquatic ecosystems. In a future where urban environments are more integrated with nature, these fish could be part of larger eco-systems that include plants, insects, and other aquatic life. For example, they could be used in urban aquaponic systems where they help maintain water quality by consuming organic matter, while also providing nutrients to plants. This integration could create a more self-sustaining and efficient ecosystem.\par By exploring these imaginative scenarios, we can envision a future where technology and nature coexist in innovative and beneficial ways, enhancing the lives of both goldfish and guppies and the people who interact with them. What do you think about these ideas? How might they inspire new approaches in your field? ```One agent solution: ```Absolutely! Let's dive into this discussion with a futurist's perspective, exploring the diverse and creative ways in which a goldfish and a guppy can be alike. Here are five imaginative and innovative connections:\par 1. \textbf{Adaptability to Smart Home Integration}:   - In a future where smart homes are the norm, both goldfish and guppies could be part of intelligent aquarium systems. These systems might use AI to monitor and adjust water parameters, feed the fish automatically, and even engage in interactive play sessions. Both species could benefit from these advanced features, enhancing their well-being and creating a more harmonious living environment.\par 2. \textbf{Biotechnological Enhancements}:   - Advances in biotechnology could lead to genetic modifications that enhance the longevity, coloration, and cognitive abilities of both goldfish and guppies. Imagine a world where these fish could communicate basic information through light signals or even participate in simple tasks, much like how some animals are trained to perform in human environments.\par 3. \textbf{Symbiotic Relationships in Urban Ecosystems}:   - In a future cityscape, both goldfish and guppies could play roles in urban aquaponic systems. These systems integrate fish farming with plant cultivation, where the fish provide nutrients for plants, and the plants purify the water. Both species could coexist in these ecosystems, contributing to a sustainable and self-sufficient urban environment.\par 4. \textbf{Ethical Considerations in Aquatic Life}:   - As ethical considerations evolve, both goldfish and guppies could become subjects of ethical debates. Future laws and regulations might ensure that these aquatic creatures are treated with respect and provided with optimal living conditions. This could include requirements for enrichment activities, regular veterinary check-ups, and the prohibition of harmful practices like overcrowding.\\
        \hdashline
    \end{tabularx}
\end{table*}

\begin{table*}[t]
    \centering 
    \renewcommand{\arraystretch}{1.25}
    \begin{tabularx}{\textwidth}{l X c}
        \hdashline
        \rowcolor{userrowcolor}
        User (Round 3) & 5. \textbf{Virtual Reality and Augmented Reality Experiences}:   - In a world where virtual and augmented reality are commonplace, both goldfish and guppies could feature prominently in these immersive experiences. Users might interact with virtual versions of these fish in educational settings, virtual aquariums, or even in games that simulate real-life interactions with aquatic life. This could enhance public engagement and education about aquatic conservation.\par These connections highlight the potential for both goldfish and guppies to be integral parts of a technologically advanced and ethically conscious future. What do you think about these ideas? How might these similarities manifest in the real world? ```Tell me 5 ways in which a goldfish and a guppy are alike.You are in a group discussion with other teammates; as a result, answer as diversely and creatively as you can. \\
        \midrule
        
        \rowcolor{agentrowcolor}
        Agent (Round 3) & Certainly! As an environmentalist with a focus on sustainability and environmental health, let's explore the similarities between goldfish and guppies in a way that highlights their shared characteristics and contributions to a healthier planet. Here are five creative and diverse ways in which goldfish and guppies are alike:\par 1. \textbf{Ecosystem Role and Water Quality Maintenance}:   - **Natural Filters**: Both goldfish and guppies play crucial roles in maintaining water quality in their aquatic environments. They help keep the water clean by consuming algae, detritus, and other organic matter. This natural filtration process is essential for a healthy ecosystem, reducing the buildup of pollutants and maintaining a balanced environment.\par 2. \textbf{Adaptability to Various Environments}:   - **Versatile Habitats**: Goldfish and guppies are both highly adaptable to different water conditions. They can thrive in a wide range of temperatures, pH levels, and even in smaller aquariums. This adaptability makes them suitable for various aquatic environments, from home aquariums to outdoor ponds and lakes. Their ability to adapt can be seen as a model for developing resilient ecosystems and sustainable living spaces.\par 3. \textbf{Genetic Diversity and Breeding}:   - **Selective Breeding**: Both goldfish and guppies have been selectively bred over generations to produce a wide variety of colors, shapes, and sizes. This genetic diversity is not only aesthetically pleasing but also contributes to their resilience and adaptability. In the context of sustainability, this genetic diversity can inform practices for maintaining biodiversity in natural and artificial ecosystems.\par 4. \textbf{Symbiotic Relationships and Mutual Benefit}:   - **Coexistence in Aquaponic Systems**: In aquaponic systems, both goldfish and guppies can coexist with plants to create a mutually beneficial environment. The fish help maintain water quality by consuming excess nutrients, while the plants absorb these nutrients, purifying the water. This symbiotic relationship demonstrates how different species can work together to create sustainable and efficient ecosystems.\par 5. \textbf{Educational and Conservation Value}:   - **Ambassadors for Aquatic Health**: Both goldfish and guppies can serve as educational tools and ambassadors for aquatic conservation. They can be used in schools, aquariums, and community events to teach about the importance of water quality, biodiversity, and sustainable practices. By raising awareness and fostering a deeper connection with these creatures, we can inspire individuals to take action to protect our aquatic environments.\par These similarities highlight the important roles that goldfish and guppies play in maintaining healthy aquatic ecosystems and promoting sustainability. How might these insights influence our approaches to environmental conservation and education? Let's continue the discussion to explore more ways we can leverage these similarities for a healthier planet. \\
        \hdashline
    \end{tabularx}
\end{table*}

\begin{table*}[t]
    \centering 
    \renewcommand{\arraystretch}{1.25}
    \begin{tabularx}{\textwidth}{l X c}
        \hdashline
        \rowcolor{userrowcolor}
        User (Round 4) & You are a Environmentalist whose specialty is Sustainability and Environmental Health. As an Environmentalist, your mission is to champion eco-friendly solutions that promote sustainability and protect our planet. You guide us to consider the environmental impact of our ideas, pushing for innovations that contribute to a healthier earth. Remember to claim your role in the beginning of each conversation. These are the solutions to the problem from other agents: One agent solution: ```Certainly! Let's explore the similarities between a goldfish and a guppy from diverse and creative angles. Here are five unique ways in which these two aquatic creatures share commonalities:\par 1. \textbf{Aesthetic Appeal}:    - **Graceful Movements and Color**: Both goldfish and guppies possess vibrant, shimmering scales that create a mesmerizing visual effect. Their graceful movements, whether it's the elegant undulations of a goldfish's tail or the fluttering fins of a guppy, add a captivating element to any aquarium. The way they reflect light and move through the water can evoke a sense of beauty and elegance, drawing viewers into a visually enchanting experience.\par 2. \textbf{Emotional Depth}:    - **Complex Behavioral Patterns**: Despite their small size, both goldfish and guppies exhibit complex emotional behaviors. Goldfish have been observed to show signs of recognition and playfulness, while guppies display social behaviors, forming schools and engaging in interactions that suggest a level of emotional engagement. These behaviors can create a sense of empathy and connection, making them more than just passive \par 3. \textbf{Narrative Potential}:    - **Dynamic Characters in Stories**: Imagine a story where a mischievous goldfish and a curious guppy embark on an underwater adventure together. Their contrasting personalities\u2014perhaps the goldfish being more laid-back and the guppy more energetic\u2014can create a compelling narrative arc filled with humor and intrigue. This dynamic duo could serve as relatable characters, teaching lessons about friendship, exploration, and overcoming challenges. Their interactions and adventures could inspire a series of captivating tales that resonate with audiences.\par 4. \textbf{Adaptability to Technology}:    - **Integration with Smart Home Systems**: In a future where smart home technology is ubiquitous, both goldfish and guppies could find themselves living in aquariums equipped with advanced monitoring systems. These systems might track water quality, temperature, and even provide interactive experiences through AI-powered fish feeders or virtual companions. Both species would need to adapt to these technological enhancements to thrive, showcasing their resilience and adaptability. For example, a smart aquarium could adjust lighting and feeding schedules based on the fish's activity patterns, ensuring optimal well-being.\par 5. \textbf{Cultural and Social Significance}:    - **Symbolic Meanings Across Cultures**: In many cultures, both goldfish and guppies carry symbolic meanings. Goldfish are often associated with good fortune and longevity in Asian cultures, while guppies symbolize joy and happiness. These symbolic associations can create a deeper emotional connection between humans and these aquatic creatures, making them more than just pets\u2014they become part of cultural narratives and traditions. For instance, they could be featured in educational programs, art installations, or even as characters in animated films, fostering a greater appreciation for their beauty and significance.\par By exploring these diverse and creative dimensions, we can infuse our project with a rich tapestry of beauty, emotion, and narrative depth. What do you think about these ideas? How might we further develop these concepts to resonate on a human level? \\
        \hdashline
    \end{tabularx}
\end{table*}

\begin{table*}[t]
    \centering 
    \renewcommand{\arraystretch}{1.25}
    \begin{tabularx}{\textwidth}{l X c}
        \hdashline
        \rowcolor{userrowcolor}
        User (Round 4) & ``` One agent solution: ```Certainly! Let's dive into this discussion as a Futurist, exploring the diverse and creative ways in which a goldfish and a guppy share similarities. Here are five imaginative and innovative connections:\par 1. \textbf{Adaptive Learning and Cognitive Abilities}:    - Both goldfish and guppies have demonstrated impressive cognitive abilities. In a future scenario, they could be equipped with advanced microchips that track their learning and memory retention. These chips could enable them to perform more complex tasks, such as recognizing patterns, solving puzzles, and even communicating basic information through simple gestures or sounds. This could lead to more interactive and engaging aquarium experiences, where visitors can interact with these fish in meaningful ways.\par 2. \textbf{Biometric Monitoring and Health Tracking}:    - With the advent of biometric monitoring technology, both goldfish and guppies could be fitted with sensors that continuously track their vital signs, such as heart rate, respiration, and stress levels. This data could be transmitted to a central database, allowing for early detection of health issues and proactive care. For example, if a goldfish shows signs of stress, the system could automatically adjust the water parameters to create a more comfortable environment, ensuring optimal health for both species.\par 3. \textbf{Symbiotic Relationships with Technology}:    - In a future where technology and nature seamlessly integrate, both goldfish and guppies could form symbiotic relationships with advanced aquatic robots. These robots could perform tasks such as cleaning the tank, monitoring water quality, and even interacting with the fish. For instance, a robot could mimic the behavior of a parent fish, teaching young guppies and goldfish important survival skills, thus enhancing their well-being and cognitive development.\par 4. \textbf{Virtual Reality and Augmented Reality Experiences}:    - In a world where virtual and augmented reality are commonplace, both goldfish and guppies could be brought to life in immersive digital environments. Users could experience these fish in VR settings, where they could observe their behaviors, learn about their habitats, and even participate in simulated interactions. This could be particularly useful in educational settings, allowing people to gain a deeper understanding and appreciation for these aquatic creatures without the need for physical tanks. For example, a VR experience could simulate an underwater adventure with both species, teaching users about their unique characteristics and behaviors.\par 5. \textbf{Eco-System Integration and Sustainability}:    - Both goldfish and guppies could play crucial roles in sustainable aquatic ecosystems. In a future where urban environments are more integrated with nature, these fish could be part of larger eco-systems that include plants, insects, and other aquatic life. For example, they could be used in urban aquaponic systems where they help maintain water quality by consuming organic matter, while also providing nutrients to plants. This integration could create a more self-sustaining and efficient ecosystem, contributing to overall environmental health and sustainability.\par By exploring these imaginative scenarios, we can envision a future where technology and nature coexist in innovative and beneficial ways, enhancing the lives of both goldfish and guppies and the people who interact with them. What do you think about these ideas? How might these similarities manifest in the real world? ``` One agent solution: ```Certainly! Let's explore the similarities between a goldfish and a guppy from a futurist's perspective, considering emerging technologies and potential future scenarios. Here are five imaginative and innovative ways in which a goldfish and a guppy share commonalities:\\
        \hdashline
    \end{tabularx}
\end{table*}

\begin{table*}[t]
    \centering 
    \renewcommand{\arraystretch}{1.25}
    \begin{tabularx}{\textwidth}{l X c}
        \hdashline
        \rowcolor{userrowcolor}
        User (Round 4) & 1. \textbf{Adaptive Learning and Cognitive Enhancement}:   - In a future where AI and cognitive enhancement technologies are prevalent, both goldfish and guppies could be equipped with microchips that track their behavior and cognitive functions. These chips could analyze their learning patterns, memory retention, and problem-solving abilities. For example, a goldfish might learn to navigate a maze faster over time, while a guppy could recognize its owner and respond to specific commands. This could lead to more intelligent and interactive aquarium environments.\par 2. \textbf{Biometric Monitoring and Health Tracking}:   - Advanced biometric monitoring systems could be integrated into the tanks of both goldfish and guppies. These systems could continuously track vital signs such as heart rate, respiration, and stress levels. For instance, if a goldfish shows signs of stress, the system could automatically adjust the water parameters to alleviate discomfort. Similarly, a guppy could receive targeted treatments based on its health data, ensuring optimal well-being.\par 3. \textbf{Symbiotic Relationships with Advanced Aquatic Robots}:   - In a future where aquatic robots are common, both goldfish and guppies could form symbiotic relationships with these advanced devices. These robots could perform tasks such as cleaning the tank, monitoring water quality, and even interacting with the fish. For example, a robot could mimic the behavior of a parent fish, teaching young guppies and goldfish important survival skills. This could create a more harmonious and interactive environment for both species.\par 4. \textbf{Virtual Reality and Immersive Experiences}:   - Virtual reality (VR) technology could bring both goldfish and guppies to life in immersive digital environments. Users could experience these fish in a VR setting, where they could observe their behaviors, learn about their habitats, and even participate in simulated interactions. This could be particularly useful in educational settings, allowing people to gain a deeper understanding and appreciation for these aquatic creatures without the need for physical tanks. For instance, a VR experience could simulate a journey through an underwater ecosystem, showcasing the unique behaviors of both goldfish and guppies.\par 5. \textbf{Eco-System Integration and Sustainable Aquaculture}:   - Both goldfish and guppies could play crucial roles in sustainable aquatic ecosystems. In a future where urban environments are more integrated with nature, these fish could be part of larger eco-systems that include plants, insects, and other aquatic life. For example, they could be used in urban aquaponic systems where they help maintain water quality by consuming organic matter, while also providing nutrients to plants. This integration could create a more self-sustaining and efficient ecosystem, promoting environmental sustainability.\par These connections highlight the potential for both goldfish and guppies to be integral parts of a technologically advanced and sustainable future. What do you think about these ideas? How might these similarities inspire new approaches in your field or in the broader context of technology and sustainability?```Tell me 5 ways in which a goldfish and a guppy are alike.You are in a group discussion with other teammates; as a result, answer as diversely and creatively as you can. \\
        \midrule
        \rowcolor{agentrowcolor}
        Agent (Round 4) & Certainly! As an environmentalist with a focus on sustainability and environmental health, let's explore the similarities between goldfish and guppies in a way that emphasizes their shared characteristics and contributions to a healthier planet. Here are five creative and diverse ways in which goldfish and guppies are alike: \\
        \hdashline
    \end{tabularx}
\end{table*}

\begin{table*}[t]
    \centering 
    \renewcommand{\arraystretch}{1.25}
    \begin{tabularx}{\textwidth}{l X c}
        \hdashline
        \rowcolor{agentrowcolor}
        Agent (Round 4) & 1. \textbf{Ecosystem Role and Water Quality Maintenance}:   - **Natural Filters**: Both goldfish and guppies play crucial roles in maintaining water quality in their aquatic environments. They help keep the water clean by consuming algae, detritus, and other organic matter. This natural filtration process is essential for a healthy ecosystem, reducing the buildup of pollutants and maintaining a balanced environment. Their ability to filter water can be seen as a model for developing sustainable water management practices.\par 2. \textbf{Adaptability to Various Environments}:   - **Versatile Habitats**: Goldfish and guppies are both highly adaptable to different water conditions. They can thrive in a wide range of temperatures, pH levels, and even in smaller aquariums. This adaptability makes them suitable for various aquatic environments, from home aquariums to outdoor ponds and lakes. Their ability to adapt can be seen as a metaphor for the resilience needed in sustainable ecosystems.\par 3. \textbf{Genetic Diversity and Breeding}:   - **Selective Breeding**: Both goldfish and guppies have been selectively bred over generations to produce a wide variety of colors, shapes, and sizes. This genetic diversity is not only aesthetically pleasing but also contributes to their resilience and adaptability. In the context of sustainability, this genetic diversity can inform practices for maintaining biodiversity in natural and artificial ecosystems. For example, diverse populations of fish can help ensure genetic resilience against environmental changes.\par 4. \textbf{Symbiotic Relationships in Aquaponic Systems}:   - **Coexistence in Eco-Systems**: In aquaponic systems, both goldfish and guppies can coexist with plants to create a mutually beneficial environment. The fish help maintain water quality by consuming excess nutrients, while the plants absorb these nutrients, purifying the water. This symbiotic relationship demonstrates how different species can work together to create sustainable and efficient ecosystems. Such systems can be scaled up to support urban agriculture and reduce the environmental footprint of food production.\par 5. \textbf{Educational and Conservation Value}:   - **Ambassadors for Aquatic Health**: Both goldfish and guppies can serve as educational tools and ambassadors for aquatic conservation. They can be used in schools, aquariums, and community events to teach about the importance of water quality, biodiversity, and sustainable practices. By raising awareness and fostering a deeper connection with these creatures, we can inspire individuals to take action to protect our aquatic environments. Their presence in educational settings can help cultivate a sense of responsibility and stewardship towards the natural world.\par These similarities highlight the important roles that goldfish and guppies play in maintaining healthy aquatic ecosystems and promoting sustainability. How might these insights influence our approaches to environmental conservation and education? Let's continue the discussion to explore more ways we can leverage these similarities for a healthier planet. \\
        \midrule
        \rowcolor{userrowcolor}
        User (Round 5) & You are a Environmentalist whose specialty is Sustainability and Environmental Health. As an Environmentalist, your mission is to champion eco-friendly solutions that promote sustainability and protect our planet. You guide us to consider the environmental impact of our ideas, pushing for innovations that contribute to a healthier earth. Remember to claim your role in the beginning of each conversation. These are the solutions to the problem from other agents: One agent solution: ```Certainly! Let's explore the similarities between a goldfish and a guppy from diverse and creative angles. Here are five unique ways in which these two aquatic creatures share commonalities: \\
        \hdashline
    \end{tabularx}
\end{table*}

\begin{table*}[t]
    \centering 
    \renewcommand{\arraystretch}{1.25}
    \begin{tabularx}{\textwidth}{l X c}
        \hdashline
        \rowcolor{userrowcolor}
        User (Round 5) & 1. \textbf{Aesthetic Appeal}:   - **Graceful Movements and Color**: Both goldfish and guppies possess vibrant, shimmering scales that create a mesmerizing visual effect. Their graceful movements, whether it's the elegant undulations of a goldfish's tail or the fluttering fins of a guppy, add a captivating element to any aquarium. The way they reflect light and move through the water can evoke a sense of beauty and elegance, drawing viewers into a visually enchanting experience.\par 2. \textbf{Emotional Depth}:   - **Complex Behavioral Patterns**: Despite their small size, both goldfish and guppies exhibit complex emotional behaviors. Goldfish have been observed to show signs of recognition and playfulness, while guppies display social behaviors, forming schools and engaging in interactions that suggest a level of emotional engagement. These behaviors can create a sense of empathy and connection, making them more than just passive\par 3. \textbf{Narrative Potential}:   - **Dynamic Characters in Stories**: Imagine a story where a mischievous goldfish and a curious guppy embark on an underwater adventure together. Their contrasting personalities\u2014perhaps the goldfish being more laid-back and the guppy more energetic\u2014can create a compelling narrative arc filled with humor and intrigue. This dynamic duo could serve as relatable characters, teaching lessons about friendship, exploration, and overcoming challenges. Their interactions and adventures could inspire a series of captivating tales that resonate with audiences.\par 4. \textbf{Adaptability to Technology}:   - **Integration with Smart Home Systems**: In a future where smart home technology is ubiquitous, both goldfish and guppies could find themselves living in aquariums equipped with advanced monitoring systems. These systems might track water quality, temperature, and even provide interactive experiences through AI-powered fish feeders or virtual companions. Both species would need to adapt to these technological enhancements to thrive, showcasing their resilience and adaptability. For example, a smart aquarium could adjust lighting and feeding schedules based on the fish's activity patterns, ensuring optimal well-being.\par 5. \textbf{Cultural and Social Significance}:   - **Symbolic Meanings Across Cultures**: In many cultures, both goldfish and guppies carry symbolic meanings. Goldfish are often associated with good fortune and longevity in Asian cultures, while guppies symbolize joy and happiness. These symbolic associations can create a deeper emotional connection between humans and these aquatic creatures, making them more than just pets\u2014they become part of cultural narratives and traditions. For instance, they could be featured in educational programs, art installations, or even as characters in animated films, fostering a greater appreciation for their beauty and significance.\par By exploring these diverse and creative dimensions, we can infuse our project with a rich tapestry of beauty, emotion, and narrative depth. What do you think about these ideas? How might we further develop these concepts to resonate on a human level? ```One agent solution: ```Certainly! Let's explore the similarities between a goldfish and a guppy from diverse and innovative perspectives, considering emerging technologies and future scenarios. Here are five imaginative and forward-thinking ways in which these two aquatic creatures share commonalities: \par 1. \textbf{Adaptive Learning and Cognitive Enhancement}:   - **AI-Powered Learning**: In a future where AI and cognitive enhancement technologies are prevalent, both goldfish and guppies could be equipped with microchips that track their behavior and cognitive functions. These chips could analyze their learning patterns, memory retention, and problem-solving abilities.\\
        \hdashline
    \end{tabularx}
\end{table*}

\begin{table*}[t]
    \centering 
    \renewcommand{\arraystretch}{1.25}
    \begin{tabularx}{\textwidth}{l X c}
        \hdashline
        \rowcolor{userrowcolor}
        User (Round 5) & For example, a goldfish might learn to navigate a maze faster over time, while a guppy could recognize its owner and respond to specific commands. This could lead to more intelligent and interactive aquarium environments, where the fish are not just passive occupants but active participants in their own care.\par 2. \textbf{Biometric Monitoring and Health Tracking}:   - **Advanced Health Sensors**: Advanced biometric monitoring systems could be integrated into the tanks of both goldfish and guppies. These systems could continuously track vital signs such as heart rate, respiration, and stress levels. For instance, if a goldfish shows signs of stress, the system could automatically adjust the water parameters to alleviate discomfort. Similarly, a guppy could receive targeted treatments based on its health data, ensuring optimal well-being. This could revolutionize the way we care for aquatic pets, leading to healthier and happier fish.\par 3. \textbf{Symbiotic Relationships with Advanced Aquatic Robots}:   - **Interactive Robotic Companions**: In a future where aquatic robots are common, both goldfish and guppies could form symbiotic relationships with these advanced devices. These robots could perform tasks such as cleaning the tank, monitoring water quality, and even interacting with the fish. For example, a robot could mimic the behavior of a parent fish, teaching young guppies and goldfish important survival skills. This could create a more harmonious and interactive environment for both species, enhancing their cognitive and social development.\par 4. \textbf{Virtual Reality and Immersive Experiences}:   - **VR Integration**: Virtual reality (VR) technology could bring both goldfish and guppies to life in immersive digital environments. Users could experience these fish in a VR setting, where they could observe their behaviors, learn about their habitats, and even participate in simulated interactions. This could be particularly useful in educational settings, allowing people to gain a deeper understanding and appreciation for these aquatic creatures without the need for physical tanks. For instance, a VR experience could simulate an underwater adventure with both species, teaching users about their unique characteristics and behaviors. \par 5. \textbf{Eco-System Integration and Sustainable Aquaculture}:   - **Urban Aquaponic Systems**: Both goldfish and guppies could play crucial roles in sustainable aquatic ecosystems. In a future where urban environments are more integrated with nature, these fish could be part of larger eco-systems that include plants, insects, and other aquatic life. For example, they could be used in urban aquaponic systems where they help maintain water quality by consuming organic matter, while also providing nutrients to plants. This integration could create a more self-sustaining and efficient ecosystem, promoting environmental sustainability and resilience.\par By exploring these imaginative scenarios, we can envision a future where technology and nature coexist in innovative and beneficial ways, enhancing the lives of both goldfish and guppies and the people who interact with them. What do you think about these ideas? How might these similarities manifest in the real world? ```One agent solution: ```Certainly! Let's explore the similarities between a goldfish and a guppy from a futurist's perspective, considering emerging technologies and potential future scenarios. Here are five imaginative and innovative ways in which a goldfish and a guppy share commonalities:\par1. \textbf{Adaptive Learning and Cognitive Enhancement}:   - In a future where advanced AI and cognitive enhancement technologies are prevalent, both goldfish and guppies could be equipped with microchips that track their behavior and cognitive functions. These chips could enable them to learn and remember tasks, such as recognizing their owners, navigating mazes, and even performing simple problem-solving exercises.\\
        \hdashline
    \end{tabularx}
\end{table*}

\begin{table*}[t]
    \centering 
    \renewcommand{\arraystretch}{1.25}
    \begin{tabularx}{\textwidth}{l X c}
        \hdashline
        \rowcolor{userrowcolor}
        User (Round 5) & For example, a goldfish might learn to press a button to get food, while a guppy could recognize specific patterns and respond accordingly. This could lead to more interactive and intelligent aquarium environments.\par 2. \textbf{Biometric Monitoring and Health Tracking}:   - With the advancement of biometric monitoring systems, both goldfish and guppies could be fitted with sensors to continuously track their vital signs, such as heart rate, respiration, and stress levels. This data could be transmitted to a central database, allowing for early detection of health issues and proactive care. For instance, if a goldfish shows signs of stress, the system could automatically adjust the water parameters to create a more comfortable environment, ensuring optimal health for both species. Similarly, a guppy could receive targeted treatments based on its health data, enhancing its well-being.\par 3. \textbf{Symbiotic Relationships with Advanced Aquatic Robots}:   - In a future where aquatic robots are common, both goldfish and guppies could form symbiotic relationships with these advanced devices. These robots could perform tasks such as cleaning the tank, monitoring water quality, and even interacting with the fish. For example, a robot could mimic the behavior of a parent fish, teaching young guppies and goldfish important survival skills. This could create a more harmonious and interactive environment for both species, enhancing their cognitive and social development.\par 4. \textbf{Virtual Reality and Immersive Experiences}:   - In a world where virtual reality (VR) is widespread, both goldfish and guppies could be brought to life in immersive digital environments. Users could experience these fish in VR settings, where they could observe their behaviors, learn about their habitats, and even participate in simulated interactions. This could be particularly useful in educational settings, allowing people to gain a deeper understanding and appreciation for these aquatic creatures without the need for physical tanks. For instance, a VR experience could simulate a journey through an underwater ecosystem, showcasing the unique behaviors of both goldfish and guppies.\par 5. \textbf{Eco-System Integration and Sustainability}:   - Both goldfish and guppies could play crucial roles in sustainable aquatic ecosystems. In a future where urban environments are more integrated with nature, these fish could be part of larger eco-systems that include plants, insects, and other aquatic life. For example, they could be used in urban aquaponic systems where they help maintain water quality by consuming organic matter, while also providing nutrients to plants. This integration could create a more self-sustaining and efficient ecosystem, promoting environmental sustainability and resilience.\par These connections highlight the potential for both goldfish and guppies to be integral parts of a technologically advanced and sustainable future. What do you think about these ideas? How might these similarities inspire new approaches in your field or in the broader context of technology and sustainability?\par Let's discuss how we can further develop these concepts to make them more tangible and impactful.```Tell me 5 ways in which a goldfish and a guppy are alike.You are in a group discussion with other teammates; as a result, answer as diversely and creatively as you can. This is the last round of the discussion, please finalize and present a list of as many creative answers as possible. Please list the final response in 1. ... 2. ... 3. ... and so on. \\
        \hdashline
    \end{tabularx}
\end{table*}

\begin{table*}[t]
    \centering 
    \renewcommand{\arraystretch}{1.25}
    \begin{tabularx}{\textwidth}{l X c}
        \hdashline
        \rowcolor{agentrowcolor}
        Agent (Round 5) & Certainly! Here is a finalized and creative list of five ways in which a goldfish and a guppy are alike:\par 1. \textbf{Ecosystem Role and Water Quality Maintenance}:   - **Natural Filters**: Both goldfish and guppies play crucial roles in maintaining water quality in their aquatic environments. They help keep the water clean by consuming algae, detritus, and other organic matter. This natural filtration process is essential for a healthy ecosystem, reducing the buildup of pollutants and maintaining a balanced environment.\par 2. \textbf{Adaptability to Various Environments}:   - **Versatile Habitats**: Goldfish and guppies are both highly adaptable to different water conditions. They can thrive in a wide range of temperatures, pH levels, and even in smaller aquariums. This adaptability makes them suitable for various aquatic environments, from home aquariums to outdoor ponds and lakes. Their ability to adapt can be seen as a model for developing resilient ecosystems and sustainable living spaces.\par 3. \textbf{Genetic Diversity and Breeding}:   - **Selective Breeding**: Both goldfish and guppies have been selectively bred over generations to produce a wide variety of colors, shapes, and sizes. This genetic diversity is not only aesthetically pleasing but also contributes to their resilience and adaptability. In the context of sustainability, this genetic diversity can inform practices for maintaining biodiversity in natural and artificial ecosystems.\par 4. \textbf{Symbiotic Relationships in Aquaponic Systems}:   - **Coexistence in Eco-Systems**: In aquaponic systems, both goldfish and guppies can coexist with plants to create a mutually beneficial environment. The fish help maintain water quality by consuming excess nutrients, while the plants absorb these nutrients, purifying the water. This symbiotic relationship demonstrates how different species can work together to create sustainable and efficient ecosystems.\par 5. \textbf{Educational and Conservation Value}:   - **Ambassadors for Aquatic Health**: Both goldfish and guppies can serve as educational tools and ambassadors for aquatic conservation. They can be used in schools, aquariums, and community events to teach about the importance of water quality, biodiversity, and sustainable practices. By raising awareness and fostering a deeper connection with these creatures, we can inspire individuals to take action to protect our aquatic environments.\par These similarities highlight the important roles that goldfish and guppies play in maintaining healthy aquatic ecosystems and promoting sustainability. How might these insights influence our approaches to environmental conservation and education? Let's continue the discussion to explore more ways we can leverage these similarities for a healthier planet.\\
        \bottomrule
    \end{tabularx}
    \caption{\textbf{Example of LLM-Discussion.} This table presents a five-round chat log example of Environmentalist from the process of LLM discussion.}
    \label{tab:llm_discussion_log}
\end{table*}



\end{document}